\documentclass{egpubl}
\usepackage[T1]{fontenc}
\usepackage{dfadobe}  

\usepackage{cite}  
\BibtexOrBiblatex
\electronicVersion
\PrintedOrElectronic
\ifpdf \usepackage[pdftex]{graphicx} \pdfcompresslevel=9
\else \usepackage[dvips]{graphicx} \fi

\usepackage{egweblnk}

\usepackage{times}
\usepackage{epsfig}
\usepackage{graphicx}
\usepackage{amsmath}
\usepackage{amssymb}
\usepackage[table,xcdraw]{xcolor}
\usepackage{colortbl}
\definecolor{Green}{rgb}{0.196, 0.854, 0.186}
\definecolor{PineGreen}{rgb}{0.561,0.0,1.0}
\definecolor{Yellow}{rgb}{1.0, 1.0, 0.0}
\definecolor{Blue}{rgb}{0.0, 0.0, 1.0}
\usepackage{tikz}
\usepackage{multirow, multicol}
\usepackage{comment}
\usepackage{pifont}
\usepackage{svg}

\title[SS-SfP: Neural Inverse Rendering for Self Supervised Shape from (Mixed) Polarization]%
{SS-SfP: Neural Inverse Rendering for Self Supervised Shape from (Mixed) Polarization}

\author[Ashish Tiwari \& Shanmuganathan Raman]
{\parbox{\textwidth}{\centering Ashish Tiwari\orcid{0000-0002-4462-6086} and Shanmuganathan Raman\orcid{0000-0003-2718-7891} 
	}
	\\
	{\parbox{\textwidth}{\centering Computer Vision, Imaging and, Graphics Lab, Indian Institute of Technology Gandhinagar, India
		}
	}
}

\begin{document}
	
	
	\maketitle
	\begin{abstract}
		We present a novel inverse rendering-based framework to estimate the 3D shape (per-pixel surface normals and depth) of objects and scenes from single-view polarization images, the problem popularly known as Shape from Polarization (SfP). The existing physics-based and learning-based methods for SfP perform under certain restrictions, i.e., (a) purely diffuse or purely specular reflections, which are seldom in the real surfaces, (b) availability of the ground truth surface normals for direct supervision that are hard to acquire and are limited by the scanner's resolution, and (c) known refractive index. To overcome these restrictions, we start by learning to separate the partially-polarized diffuse and specular reflection components, which we call reflectance cues, based on a modified polarization reflection model and then estimate shape under mixed polarization through an inverse-rendering based self-supervised deep learning framework called SS-SfP, guided by the polarization data and estimated reflectance cues. Furthermore, we also obtain the refractive index as a non-linear least squares solution. Through extensive quantitative and qualitative evaluation, we establish the efficacy of the proposed framework over simple single-object scenes from DeepSfP dataset and complex in-the-wild scenes from SPW dataset in an entirely self-supervised setting. To the best of our knowledge, this is the first learning-based approach to address SfP under mixed polarization in a completely self-supervised framework. 


	\end{abstract}  
	\begin{figure*}[t]
		\centering
		\includegraphics[width=0.95\textwidth]{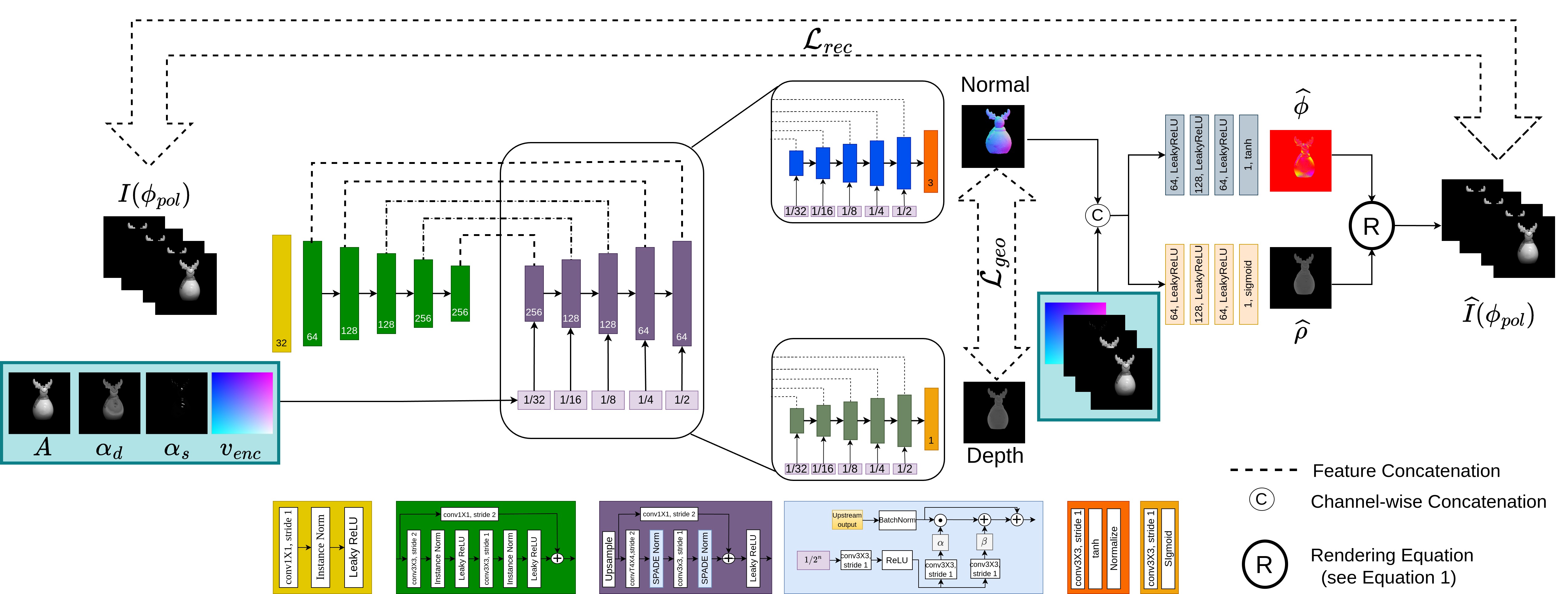}
		\caption{Layer-wise detailed description of the proposed framework - SS-SfP.}
		\label{fig:net_arch}
		\vspace{-0.50cm}
	\end{figure*}
	
	\section{Introduction}
	\label{sec:intro}
	The polarization state of the light reflected off a surface depends on the shape of the underlying surface, which is the basic principle of the Shape from Polarization (SfP) problem. The prime objective is to estimate surface normals from a set of polarization images. The physics involved in SfP arises from Fresnel Equations \cite{collett2005field} and is among the most complex problems in addressing computer vision tasks such as 3D reconstruction. Although several attempts have addressed SfP by providing physics-based solutions, researchers have only recently started looking at SfP through the lens of deep learning \cite{ba2020deep, lei2022shape}. However, since a physics-only or a learning-only solution fails to achieve the best performance individually, Ba \emph{et al.} \cite{ba2020deep} propose combining physical priors with a deep network to address the SfP problem. While other photometric methods for 3D reconstruction, such as Photometric Stereo (PS) \cite{woodham1980photometric} and Shape from Shading (SfS) \cite{prados2005shape} require known or estimated lighting directions for shape estimation, SfP can be used under completely passive lighting conditions without knowing lighting directions. Consider, for example, a color poster on a wall. Unlike the RGB-based methods, such as PS \cite{woodham1980photometric}, SfP does not get distracted by mere image semantics (of the poster) to provide erroneous surface normal estimates of the underlying surface (the wall).  The polarization cues reflect the changes in the polarization state of the reflected light instead of just the RGB measurement. Further, it applies to a broad category of materials \cite{mingqi2022transparent} such as translucent, transparent, dielectrics, and metals with applications such as image segmentation \cite{shabayek2012vision}, robot navigation \cite{berger2017depth}, image enhancement \cite{schechner2011inversion}, and underwater imaging \cite{schechner2015self}. The underlying Fresnel equations \cite{collett2005field} in modeling SfP pose a different set of ambiguities arising due to model mismatch and periodicity in the sinusoidal behaviour of the reflected polarized light intensity with respect to the polarizer angle (detailed in the supplementary). While researchers have used additional information such as shading constraints, surface convexity, and coarse depth maps to address such ambiguities, they have not succeeded sufficiently \cite{atkinson2007shape,kadambi2015polarized}. The requirement of a known refractive index brings another challenge as it introduces \textit{refractive distortion} in the estimated shape if the refractive index is incorrect. The assumption of the diffuse-only or specular-only reflections from surface points is highly restrictive. A broad class of methods (discussed in Section \ref{sec:rw}) use computational and optical techniques to split the image into specular-only or diffuse-only images. However, in the real world, surfaces are neither purely diffuse nor purely specular but somewhere in-between, i.e., they tend to exhibit mixed polarization where the camera receives both diffusely and specularly reflected light. The existing reflectance decomposition techniques (specific to the SfP setting) have primarily relied on image intensity or color information \cite{nishino2001determining,ma2007rapid, lin2002diffuse, tan2008separating}, making them susceptible to artefacts. While some have used polarization cues, they needed active illumination to perform the separation \cite{ma2007rapid, ghosh2011multiview}.\\
	\indent\textbf{Contributions.} The following is a summary of the key contributions of the work.\vspace{-0.2cm}
	\begin{itemize}
		\item We analyze the polarization image formation model under mixed reflection and estimate diffuse and specular reflection components at each pixel, calling them the \textit{reflectance cues}, using the least-squares approach.
		\item We propose a neural inverse rendering based framework, called SS-SfP - \textbf{S}elf \textbf{S}upervised \textbf{S}hape \textbf{f}rom (Mixed) \textbf{P}olarization, to estimate the shape (per-pixel surface normal and depth) over surfaces exhibiting mixed polarization using the raw polarization images and reflectance cues. Further, we also obtain refractive index through a non-linear least squares formulation.
		\item To the best of our knowledge, we are the first to address the problem of Shape from (Mixed) Polarization using a deep learning framework under a self-supervised setting.
	\end{itemize}

	
	\section{Related Work}
	\label{sec:rw} In this section, we shall review some of the physics-only and learning-based SfP methods, including methods for reflectance separation.\\
	\indent \textbf{Shape from Polarization.} Several traditional physics-based methods have used a variety of cues such as coarse shading from two views \cite{atkinson2007shape}, depth maps \cite{kadambi2015polarized}, reciprocal image pairs \cite{ding2021polarimetric}, two-view stereo \cite{fukao2021polarimetric, zhu2019depth}, multi-view stereo \cite{cui2017polarimetric, miyazaki2016surface}, or front-flash illumination \cite{deschaintre2021deep}, active lighting \cite{morel2006active} to resolve the underlying ambiguities in SfP. Others have combined photometric stereo \cite{woodham1980photometric} and shading information \cite{smith2018height, mahmoud2012direct} with SfP. Baek \emph{et al.} \cite{baek2018simultaneous} choose to perform joint optimization of appearance, normals, and refractive index. Yu \emph{et al.} \cite{yu2017shape} directly estimate height from the polarization images through nonlinear least squares formulation. Moreover, Mecca \emph{et al.} \cite{mecca2017differential, logothetis2019differential} propose differential level-set-based geometric characterizations to address SfP from two-light polarimetric imaging. Guarnera \emph{et al.} \cite{guarnera2012estimating} have addressed SfP under a single spherical incident lighting condition that is either unpolarized or circularly polarized by considering the specularly reflected light. Further, other methods have used deep networks to disambiguate SfP. While Ba \emph{et al.} \cite{ba2020deep} train a CNN to obtain normals from polarization over a real-world object-level dataset, Kondo \emph{et al.} \cite{kondo2020accurate} have done the same over a synthetic dataset of polarization images with a new polarimetric BRDF model. Further, Lei \emph{et al.} \cite{lei2022shape} study scene-level SfP over complex scenes in the wild. Recently, Muglikar \emph{et al.} \cite{muglikar2023event} approached SfP through event cameras instead of conventional RGB cameras.\\
	\vspace{-0.05cm}
	\indent\textbf{Reflectance separation.} Early SfP methods have assumed pure reflection types and materials to constrain the problem. For instance, Rahmann \emph{et al.} \cite{rahmann2001reconstruction} assume pure specular reflection, and others \cite{miyazaki2003polarization, atkinson2017polarisation} assume pure diffuse reflection. Other methods perform reflectance separation using only image intensity \cite{nishino2001determining} or color information \cite{ma2007rapid, lin2002diffuse, tan2008separating} and hence, are susceptible to artefacts. Another set of approaches to separate diffuse and specular components using polarization rely on active illumination such as spherical gradient illumination \cite{ma2007rapid, ghosh2011multiview} to achieve photorealistic reconstructions. Under passive illumination, Nayar \emph{et al.} \cite{nayar1997separation} introduce a separation technique using polarization images and color cues limited by the smoothness assumption. Huynh \emph{et al.} \cite{huynh2010shape} enforce a generative model on the material dispersion equations to obtain refractive index. Taamazyan \emph{et al.} \cite{taamazyan2016shape} propose a joint optimization method to estimate shape, perform diffuse-specular separation, and per-pixel refractive indices. Ghosh \emph{et al.} \cite{ghosh2010circularly} achieve the same by analyzing Stokes reflectance field of circularly polarized spherical illumination. Recently, Dave \emph{et al.} \cite{dave2022pandora} used a learning-based framework to achieve the same except the refractive index estimation. They additionally estimate the illumination incident on the objects using polarization cues and perform the multi-view 3D reconstruction. However, our work primarily focuses on single-view reconstruction. Furthermore, Kajiyama \emph{et al.} \cite{kajiyama2023separating} achieve reflectance separation based on the polarization and dichromatic reflection models. However, they consider three-channel (RGB) color polarization images leading to twelve radiance values per pixel, unlike our approach that considers single-channel images with just four radiance values per pixel.
	\section{Shape from Polarization: Background} \label{sec:prelims}
	The polarization data can be obtained by physically rotating a linear polarizer at different angles in front of the camera. Nowadays, it is achieved through commercially available polarization cameras (such as PHX05XS-P) at four angles $\phi_{pol} = \{0, \frac{\pi}{4}, \frac{\pi}{2}, \frac{3\pi}{4}\}$ in a single shot. The irradiance $I(\phi_{pol})$ measured by the polarization camera after an incident unpolarized light is reflected from a single scene point shows a sinusoidal variation with respect to the rotation angle of the polarizer, as described in Equation \ref{eq:1}.
	\begin{equation}
	\centering
	I(\phi_{pol}) = \text{A} + \text{B} \cos(2\phi_{pol}-2\phi)
	\label{eq:1}
	\end{equation}
	Here, $A = \frac{I_{max} + I_{min}}{2}$ and $B = \frac{I_{max} - I_{min}}{2}$ with $I_{max}$ and $I_{min}$ being the maximum and minimum intensity values (respectively) obtained by rotating the polarizer at different angles. One can easily think of $A$ as the DC component and $B$ as the amplitude of the reflected polarized light. Three physical quantities - the unpolarized light intensity $(A)$, the phase angle $(\phi)$, and the degree of polarization $(\rho = \frac{B}{A})$ describe entirely the polarization state of the light. The three unknowns $(A, \rho, \phi)$ can be obtained by sampling $I(\phi_{pol})$ at minimum three different values of $\phi_{pol}$. We delve deeper into this model to discuss a more general case for mixed polarization in Section \ref{sec:method}. The surface normal parametrized by the azimuth angle $(\varphi)$ and the zenith angle $(\theta)$ at any surface point is given as follows.
	\begin{equation}
	\centering
	\mathbf{n} = \begin{bmatrix} n_{x} & n_{y} & n_{z} \end{bmatrix}^{T} = \begin{bmatrix} \sin\theta \cos\varphi  & \sin\theta \sin\varphi & \cos\theta\end{bmatrix}^{T}
	\label{eq:2}
	\end{equation}

	\section{Method}
	\label{sec:method}
	
	In this work, we aim to estimate the 3D shape (per-pixel surface normals and depth) of an object or a scene solely from single-view polarization images - the problem popularly known as Shape from Polarization (SfP). The key reason for using a deep learning framework is to overcome the inherent ambiguities and compensate for limitations in SfP by modeling the physics of polarization through a completely data-driven approach.
	The learning-based SfP methods such as \cite{ba2020deep, lei2022shape} thus far have shown improved performance over the classical physics-based methods. Interestingly, while Ba \emph{et al.} \cite{ba2020deep} show that a deep network guided by physics-based priors on surface normals has been successful in obtaining optimal solutions from the polarization information, Lei \emph{et al.} \cite{lei2022shape} have shown improved performance without including any such physics-based priors. However, both of these methods have not explicitly handled mixed reflections. Moreover, since the physical priors on surface normals solely rely on the kind of reflection, the fundamental task is to estimate somehow the nature and the extent of reflection (diffuse or specular) at each pixel.
	\vspace{-0.25cm}
	\subsection{Overview} \label{sec:overview}
	To address this, we first describe the modified polarization image formation under mixed polarization to obtain the reflectance components - diffuse component ($A_{d}$) and specular component ($A_{s}$) along with the respective degrees of polarization ($\rho_{d}$ and $\rho_{s}$) and the respective phase angles ($\phi_{d}$ and $\phi_{s}$). Next, we use an encoder-decoder-based architecture to estimate the surface normals and depth (through two separate decoder branches) using the polarization images, the reflectance cues ($\alpha_{d} = A_{d}/A$ and $\alpha_{s}=A_{s}/A$), and the view encoding ($v_{enc}$). We think of $\alpha_{d}$ and $\alpha_{s}$ as the extent of diffuse and specular reflections at each pixel. The existing methods use polarization information ($A, \rho, \phi$) along with the raw polarization images $\{I(\phi_{pol}^{i})\}$ where $i\in \{1,2,3,4\}$ as input which is seemingly redundant, since the polarization information can be derived from the raw polarization images, and can then regress to surface normals $\mathbf{\widehat{n}}$. Unlike the method by Ba \emph{et al.} \cite{ba2020deep} that uses physics-based priors on normals, we use the derived reflectance cues to guide the network for shape estimation. These physics-based priors are often ambiguous and time-consuming to compute ($\sim 1.2$ seconds, see discussion in Section \ref{sec:ablation} and Table \ref{tab:2}). View encoding accounts for non-orthographic projections for scene-level SfP, as described in \cite{lei2022shape}. Further, we use these learned surface normals (and depth derivatives) to recover the complete polarization information \{$I(\phi_{pol}), \rho, \phi$\} under mixed polarization, as per Equation \ref{eq:1}. It is to be noted that for the latter to be recovered correctly, we expect the surface normal estimates to be unambiguous and error-free. In due course, the network learns the surface properties (reflectance and refractive index) under mixed polarization by attempting to handle the underlying ambiguities in a completely self-supervised setting.

	\begin{figure}[t]
		\centering
		\includegraphics[width=\linewidth]{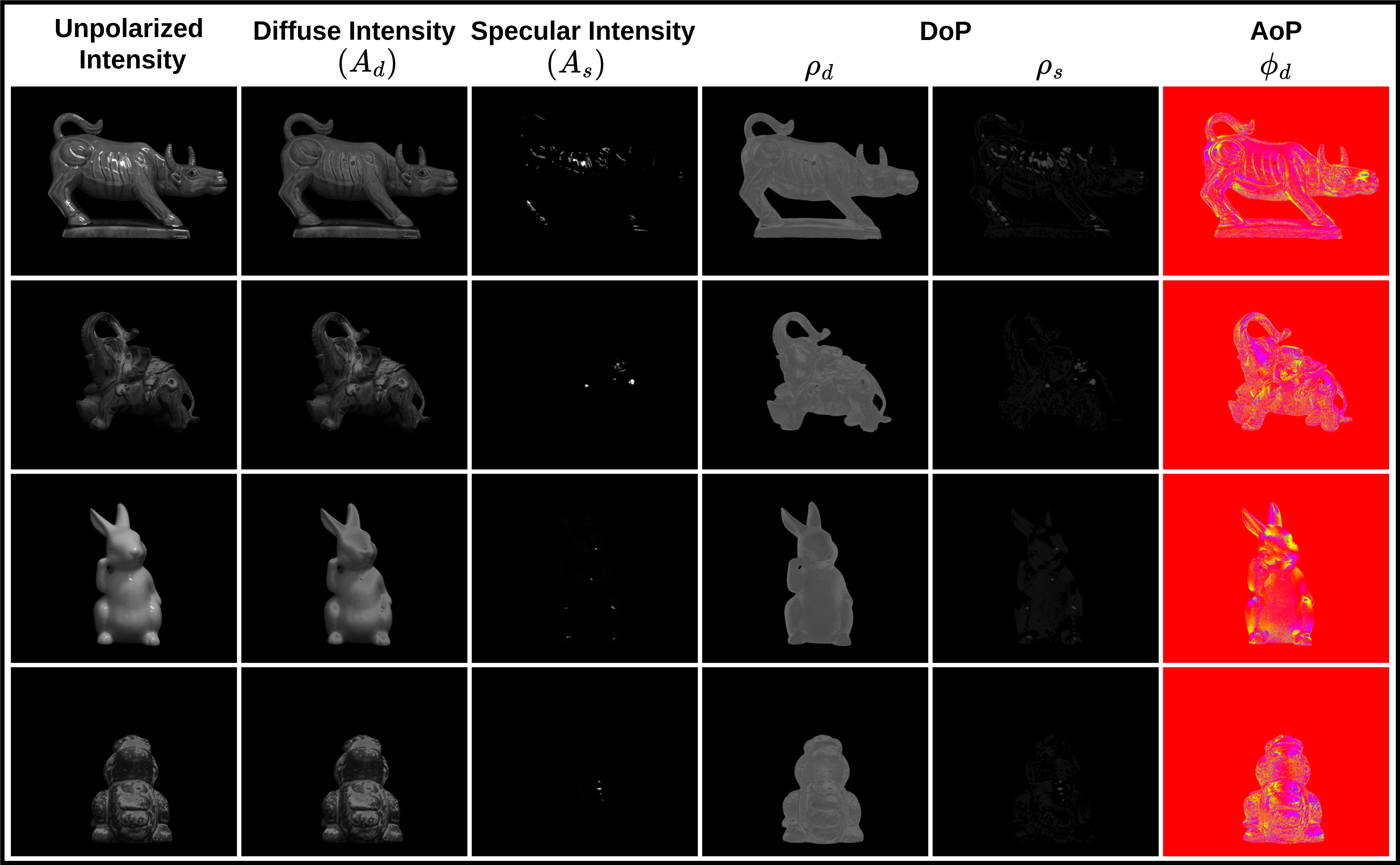}
		\caption{Qualitative results reflectance separation $(A_{d}, A_{s})$ and recovered polarization information - Angle of Polarization ($\phi_{d}$) and Degree of Polarization ($\rho_{d}, \rho_{s}$).}
		\label{fig:reflection}
		\vspace{-0.5cm}
	\end{figure}
	
	\subsection{Reflection Separation under Mixed Polarization} \label{sec:image_model}
	Consider an object or a scene illuminated by an unpolarized light source. The reflected light observed on the object/scene surface consists of partially polarized diffuse and specular reflection components. Following Equation \ref{eq:1}, the intensity of the partially polarized diffuse reflection component can be obtained as per Equation \ref{eq:A}.
	\begin{equation}
	\centering
	I_{d}(\phi_{pol}) = A_{d} + B_{d}cos(2\phi_{pol}-2\phi_{d})
	\label{eq:A}
	\end{equation}
	As discussed earlier, $A_{d}$ and $B_{d}$ are the DC component and amplitude of the diffuse reflection component, respectively. $\phi_{d}$ and $\rho_{d} = B_{d}/A_{d}$ are the phase angle and degree of polarization for the diffuse component, respectively. Along similar lines, we describe the intensity of the partially polarized specular reflection with $A_{s}$, $B_{s}$ $\phi_{s}$, and $\rho_{s} = B_{s}/A_{s}$ being DC component, the amplitude, the phase angle, and degree of polarization of the specular reflection component, respectively, as described in Equation \ref{eq:B}.
	\begin{equation}
	\centering
	I_{s}(\phi_{pol}) = A_{s} + B_{s}cos(2\phi_{pol}-2\phi_{s})
	\label{eq:B}
	\end{equation}
	Furthermore, the phase angles of diffuse and specular reflection components are related as per Equation \ref{eq:C}.
	\begin{equation}
	\centering
	\phi_{s} = \phi_{d} \pm \frac{\pi}{2}
	\label{eq:C}
	\end{equation}
	Finally, combining the polarimetric properties in Equation \ref{eq:A}, \ref{eq:B}, and \ref{eq:C} such that $I(\phi_{pol}) = I_{d}(\phi_{pol}) + I_{s}(\phi_{pol})$ \cite{taamazyan2016shape}, we can represent the radiance at a pixel for a particular polarizer angle $(\phi_{pol})$ as described in Equation \ref{eq:D}.
	\begin{equation}
	\centering
	\begin{split}
	I(\phi_{pol}) & = [A_{d} + B_{d}cos(2\phi_{pol}-2\phi_{d})] \\ & +[A_{s} - B_{s}cos(2\phi_{pol}-2\phi_{d})] 
	\end{split}
	\label{eq:D}
	\end{equation}
	Therefore, we can rewrite Equation \ref{eq:D} to obtain Equation \ref{eq:E} as follows.
	\begin{equation}
	\centering    
	\begin{split}
	I(\phi_{pol}) & = (A_{d} + A_{s}) + (B_{d} - B_{s})cos(2\phi_{pol}-2\phi_{d}) \\
	& = A_{m} + B_{m}cos(2\phi_{pol}-2\phi_{d})
	\end{split}        
	\label{eq:E}
	\end{equation}
	Here, $A_{m}$ and $B_{m}$ are the DC component and amplitude of the reflected polarized light under mixed polarization, respectively, such that the Equation \ref{eq:F1} and \ref{eq:F2} holds.
	\begin{equation}
	\centering
	A_{m} = A_{d} + A_{s}
	\label{eq:F1}
	\end{equation}
	\vspace{-0.75cm}
	\begin{equation}
	\centering
	B_{m} = B_{d} - B_{s} = \rho_{d}A_{d} - \rho_{s}A_{s} 
	\label{eq:F2}
	\end{equation}
	
	\subsubsection{Estimating $A_{m}$, $B_{m}$, and $\phi_{d}$}
	We fit a cosine curve to the pixel values observed through four polarization angles, e.g. $0^{\circ}, 45^{\circ}, 90^{\circ}$, and $135^{\circ}$ to obtain $A_{m}$, $B_{m}$, and $\phi_{d}$ (upto the $\pi$-ambiguity) as the least squares solution to a system of linear equations. This approach better incorporates robustness to noise \cite{huynh2010shape}. It is worth noting that the phase angle estimation in the case of color polarization images (as in \cite{kajiyama2023separating}) becomes a little more tedious since each color channel could have different phase angles due to the noise present in pixel values. Rewriting Equation \ref{eq:E} in the vectorized form we obtain Equation \ref{eq:7}.
	\begin{equation}
	\centering
	I(\phi_{pol}^{i}) = \begin{bmatrix}1 \\ \cos(2\phi_{pol}^{i}) \\ \sin(2\phi_{pol}^{i})  \end{bmatrix} ^{T}\begin{bmatrix}A_{m} \\ B_{m}\cos(2\phi_{d}) \\B_{m}\sin(2\phi_{d}) \end{bmatrix} 
	= f_{i}^{T}x
	\label{eq:7}
	\end{equation}
	Using this, we obtain the linear system of equations in Equation \ref{eq:8}.
	\begin{equation} \mathbb{I}=\mathbb{A}\mathbf{x} \label{eq:8}\end{equation} 
	such that $\mathbb{I} = \begin{bmatrix} I(0) \\ I(\frac{\pi}{4}) \\ I(\frac{\pi}{2}) \\ I(\frac{3\pi}{4})\end{bmatrix}$ and $\mathbb{A} = \begin{bmatrix} f_{1}^{T} \\  f_{2}^{T} \\ f_{3}^{T} \\ f_{4}^{T}\end{bmatrix}$.\\
	Solving Equation \ref{eq:8}, we get $\mathbf{x} = \begin{bmatrix} x_{1} & x_{2} & x_{3} \end{bmatrix}^{T}$ such that,
	\begin{align}
	\centering
	A_{m} = x_{1}, \hspace{0.1cm} B_{m} = \sqrt{x_{2}^{2} + x_{3}^{2}}, \hspace{0.1cm} \text{and} \hspace{0.1cm} \phi_{d} = \frac{1}{2}\tan^{-1}\Big(\frac{x_{3}}{x_{2}}\Big)
	\label{eq:9}
	\end{align}
	As per Equation \ref{eq:C}, we can obtain $\phi_{s} = \phi_{d} \pm \frac{\pi}{2}$.
	
	\subsubsection{Estimating $A_{d}$, $A_{s}$, $\rho_{d}$, and $\rho_{s}$}
	Given that we have the values of $A_{m}$ and $B_{m}$ for each pixel, we solve for the other unknowns ($A_{d}$, $A_{s}$, $\rho_{d}$, and $\rho_{s}$) in Equation \ref{eq:F1} and \ref{eq:F2} by nonlinear minimization as described in Equation \ref{eq:G}.
	\begin{equation}
	\centering
	\begin{split}
	\min_{A_{d}, A_{s}, \rho_{d},\rho_{s}} \sum_{\text{all pixels}} \Big[ [A_{m} - (A_{d} + A_{s})]^{2} + [B_{m} - (\rho_{d}A_{d} - \rho_{s}A_{s}]^{2}\Big]
	\end{split}
	\label{eq:G}
	\end{equation}
	We solve Equation \ref{eq:G} via alternating least squares estimation. We first compute $A_{d}$ and $A_{S}$ by solving Equation \ref{eq:F1} through the least squares. Here, we impose the non-negative constraints $A_{d} \geq 0$ and $A_{s} \geq 0$ since the DC components of the diffuse and specular reflection components are non-negative. Similarly, we then compute $B_{d}$ and $B_{s}$ in Equation \ref{eq:F2} through the least squares estimation under non-negative constraints ($B_{d} \geq 0$ and $B_{s} \geq 0$). However, since the phase angle is estimated with $\pi$-ambiguity, we solve the modified version of Equation \ref{eq:F2} such that we get Equation \ref{eq:H},
	\begin{equation}
	\centering
	\pm B_{m} = B_{d}-B_{s}
	\label{eq:H}
	\end{equation}
	Finally, we obtain $\rho_{d} = B_{d}/A_{d}$ and $\rho_{s} = B_{s}/A_{s}$. Figure \ref{fig:reflection} shows the diffuse and specular reflectance components for a few objects from the DeepSfP dataset \cite{ba2020deep}.
	
	\begin{table*}[t]
		\setlength{\tabcolsep}{1.5pt}
		\centering
		\resizebox{\linewidth}{!}{%
			\begin{tabular}{c|ccc|ccc|ccc}\hline
				\multirow{2}{*}{Datasets} & \multirow{2}{*}{\begin{tabular}[c]{@{}c@{}}Miyazaki \\ \emph{et al.} \cite{miyazaki2003polarization}\end{tabular}} & \multirow{2}{*}{\begin{tabular}[c]{@{}c@{}}Mahmoud\\ \emph{et al.} \cite{mahmoud2012direct}\end{tabular}} & \multirow{2}{*}{\begin{tabular}[c]{@{}c@{}}Smith\\ \emph{et al.} \cite{smith2018height}\end{tabular}} & \multicolumn{3}{c|}{Supervised (S)} & \multicolumn{3}{c}{Self-Supervised (SS)} \\
				&  &  &  & Ba \emph{et al.} \cite{ba2020deep} & Lei \emph{et al.} \cite{lei2022shape} & SS-SfP & Ba \emph{et al.} \cite{ba2020deep} & Lei \emph{et al.} \cite{lei2022shape} & SS-SfP \\ \hline
				DeepSfP & 43.94  & 51.79 & 45.39 & 18.52 & \cellcolor{Yellow}{14.68} & \cellcolor{Green}{14.61} & 27.74  & \cellcolor{Yellow}{21.59} & \cellcolor{Green}{16.89} \\
				SPW & 55.34  & 52.14 & 50.42 & 28.43 & \cellcolor{Green}{17.86} & \cellcolor{Yellow}{18.69} & 32.69 & \cellcolor{Yellow}{29.81} & \cellcolor{Green}{19.77}\\\hline
			\end{tabular}%
		}
		\caption{Quantitative comparison of SS-SfP in terms of mean angular of the estimated surface normals (in degree) with the baseline methods both evaluated under supervised (S) and self-supervised (SS) setting. The metrics reported are over the test sets of DeepSfP \cite{ba2020deep} and SPW \cite{lei2022shape} datasets. \textcolor{Green}{GREEN} and \textcolor{Yellow}{YELLOW} represent the best and the second best performance, respectively.}
		\label{tab:1}
		\vspace{-0.25cm}
	\end{table*}
	
	\subsection{Network Design}
	The detailed architecture of the proposed inverse rendering-based framework - SS-SfP is shown in Figure \ref{fig:net_arch}. We employ the widely adopted encoder-decoder architecture similar to the existing learning-based works \cite{deschaintre2021deep, ba2020deep, lei2022shape}. The encoder comprises five convolutional residual blocks, each with two convolutional layers. We use instance normalization in the encoder layers for better convergence (empirical evaluation is provided in the supplementary material). The encoder takes four polarization images $\{I(0), I(\frac{\pi}{4}), I(\frac{\pi}{2}), I(\frac{3\pi}{4})\}$ as input. The decoder is specialized through SPADE normalization \cite{park2019semantic} (inspired by \cite{ba2020deep}) and is split into two branches - one for surface normal and the other for depth estimation. Each decoder branch takes the same input from the encoder. Additionally, we inject the unpolarised intensity $(A)$, reflectance cues ($\alpha_{d}, \alpha_{s}$), and view encoding $(v_{enc})$ through SPADE normalization block into the decoder blocks to better model the surface properties and guide the network for shape estimation. For view encoding, we first compute the displacement field $(du, dv)$ at each pixel, representing the displacement of each pixel from the center of the image grid, and then normalize to $[-1, 1]$ such that $v_{enc} = \begin{bmatrix} du & dv & 1 \end{bmatrix}^{T}$. The estimated surface normal maps along with $A, \alpha_{d}, \alpha_{s}$, and $v_{enc}$ are then passed through two separate MLPs to recover $\phi$ and $\rho$, respectively, which are then passed through the rendering equation (Equation \ref{eq:1}) to reconstruct the input raw polarization images ($\widehat{I}(\phi_{pol}^{i})$). 
	
	\subsubsection{Refractive Index Estimation} The learning-based approach implicitly learns the material properties (such as reflectance and refractive index) from the polarization data instead of using an explicit polarisation model for $\rho$ with refractive index as a parameter. As described in Section \ref{sec:method}, we estimate the shape (surface normal and depth) in a refractive index invariant manner and obtain the zenith angle as $\theta = cos^{-1}(n_{z})$. With the zenith angle $(\theta)$, $\rho_{d}$, and $\rho_{s}$, we can obtain an optimal refractive index $(\eta_{opt})$ by solving a nonlinear least squares problem, as described in Equation \ref{eq:15}.
	\begin{align}
	\centering
	\eta_{opt} = arg\min_{\eta}\sum_{x} & \bigg(\Big|\Big|\alpha_{d}(x)\Big(\rho_{d}(x) - \widehat{\rho}_{d}(\theta(x), \eta)\Big)\Big|\Big|_{2}^{2} + \notag \\ & \Big|\Big|\alpha_{s}(x)\Big(\rho_{s}(x) - \widehat{\rho}_{s}(\theta(x), \eta)\Big)\Big|\Big|_{2}^{2}\bigg)
	\label{eq:15}
	\end{align}
	Here, $\widehat{\rho}_{d}$ and $\widehat{\rho}_{s}$ are reflection-specific DoPs for diffuse and specular polarization \cite{collett2005field} (described in the supplementary material). Interestingly, we rely on single-view polarimetric images for refractive index recovery, unlike previous attempts that have either considered multi-spectral images \cite{huynh2010shape, huynh2013shape} or two source polarimetric images \cite{tozza2021uncalibrated}. We assume a uniform refractive index over the entire object, as considered in earlier works \cite{ba2020deep, lei2022shape} and utilize the fact that the dependency of $\rho$ on $\eta$ is weak \cite{atkinson2006recovery}. Our refractive index estimate falls well within the range $1.3$ to $1.6$ (as prescribed for most dielectric materials) with a \underline{\textit{mean}} $1.486$ and a \underline{\textit{standard deviation}} $\pm 0.26$ across objects in the DeepSfP test set.
	
	\begin{table*}[t]
		\setlength{\tabcolsep}{10pt}
		\centering
		\resizebox{\linewidth}{!}{%
			\begin{tabular}{c|cccccc|c|c}\hline
				Objects & Box & Dragon &  Christmas & Flamingo & Horse & Vase & Whole Set & Time(s)\\ \hline
				Ba \emph{et al.} \cite{ba2020deep} (S) \ & 23.51  & 21.55 & 13.50 & 20.19 & 22.27 & \cellcolor{Yellow}{10.32} &  18.52 & 1.181    \\
				Lei \emph{et al.} \cite{lei2022shape} (S) & \cellcolor{Green}{16.21}  & \cellcolor{Green}{18.01} & \cellcolor{Green}{10.19} & \cellcolor{Green}{17.11} & \cellcolor{Green}{18.29} & \cellcolor{Green}{8.25 }& \cellcolor{Green}{14.68} & \cellcolor{Green}{0.153}\\
				SS-SfP & \cellcolor{Yellow}{19.57} & \cellcolor{Yellow}{19.14} & \cellcolor{Yellow}{13.29} & \cellcolor{Yellow}{18.62} & \cellcolor{Yellow}{20.05} & 10.61 & \cellcolor{Yellow}{16.89} & \cellcolor{Yellow}{0.183} \\\hline
				
			\end{tabular}%
		}
		\caption{Quantitative comparison over each object in the test sets of DeepSfP dataset \cite{ba2020deep} in terms of mean angular of the estimated surface normals (in degree). Last column: pre-processing time for polarization representation ($1024 \times 1224$) provided as input to different methods. Tested on single thread Intel Core i7 processor clocking at 3.60GHz. \textcolor{Green}{GREEN} and \textcolor{Yellow}{YELLOW} represent the best and the second best performance, respectively.}
		\label{tab:2}
		\vspace{-0.25cm}
	\end{table*}
	\begin{table*}[h]
		\resizebox{\linewidth}{!}{%
			\begin{tabular}{c|ccccc|c} \hline
				\begin{tabular}[c]{@{}c@{}}Lighting \\ conditions\end{tabular} & \begin{tabular}[c]{@{}c@{}}Miyazaki \\ \emph{et al.} \cite{miyazaki2003polarization} (S)\end{tabular} & \begin{tabular}[c]{@{}c@{}}Mahmoud \\ \emph{et al.} \cite{mahmoud2012direct} (S) \end{tabular} & \begin{tabular}[c]{@{}c@{}}Smith \\ \emph{et al.}  \cite{smith2018height} (S)\end{tabular} & Ba \emph{et al.} \cite{ba2020deep} (S) & Lei \emph{et al.}\cite{lei2022shape} (S) & SS-SfP \\ \hline
				Indoor & 47.66 & 50.06 & 38.82 & 17.97 & \cellcolor{Green}{13.71} & \cellcolor{Yellow}{15.79} \\
				Outdoor Cloudy & 43.24 & 46.76 & 36.51 & 16.36 & \cellcolor{Green}{13.07} & \cellcolor{Yellow}{15.12} \\
				Outdoor Sunny & 51.63 & 51.27 & 48.99 & 21.24 & \cellcolor{Green}{17.25} & \cellcolor{Yellow}{19.76}\\ \hline
			\end{tabular}%
		}
		\caption{Effect of varying lighting conditions over DeepSfP dataset \cite{ba2020deep} in terms of mean angular of the estimated surface normals (in degree). (S) and (SS) represent evaluation under supervised and self-supervised settings, respectively. \textcolor{Green}{GREEN} and \textcolor{Yellow}{YELLOW} represent the best and the second best performance, respectively.}
		\label{tab:3}
		\vspace{-0.5cm}
	\end{table*}
	
	\subsubsection{Optimization Objective}
	The entire network is optimized using a combination of reconstruction loss, geometric constraint, and polarization angle ratio constraint.\\
	\textbf{Reconstruction loss.} We minimize the L2-loss over input ($I(\phi_{pol}^{i})$) and recovered ($\widehat{I}(\phi_{pol}^{i})$) polarization images  and the desired ($\rho$)  and the recovered ($\widehat{\rho}$) DoP, and L1-loss over the desired ($\phi$) and the recovered ($\widehat{\phi}$) AoP .
	\begin{align}
	\centering
	\mathcal{L}_{rec} = & \hspace{0.2cm} \lambda_{1}\sum_{i=1}^{4}||\widehat{I}(\phi_{pol}^{i}) - I(\phi_{pol}^{i})||_{2}^{2} + \lambda_{2}||\widehat{\rho}- \rho||_{2}^{2} \notag \\ & + \lambda_{3}||\widehat{\phi}- \phi||_{1}    
	\label{eq:11}
	\end{align}
	Here, $\lambda_{1} = 1.0, \lambda_{2}=2.5,$ and $\lambda_{3} = 2.5$.\\
	\textbf{Geometric Constraint.} To ensure smooth gradient flow from $\mathcal{L}_{rec}$ especially when there is no direct supervision for surface normals, we deploy the geometry constraint between the estimated surface normals and the estimated depth map.
	\begin{equation}
	\mathcal{L}_{geo} = \sum_{\text{all pixels}}(1 - \mathbf{n}^{T}\mathbf{z}_{d})
	\label{eq:12}
	\end{equation}
	Here, $\mathbf{z}_{d} = \frac{\begin{bmatrix} -z_{x} & -z_{y} & 1\end{bmatrix}^{T}}{ \sqrt{{z_{x}}^{2} + {z_{y}}^{2} + 1}}$  with $z_{x} = \frac{\partial z}{\partial x}$ and $z_{y} = \frac{\partial z}{\partial y}$ being the first-order derivatives of the depth map in $x$ and $y$ directions, respectively.\\
	\textbf{Polarizer Angle Ratio Constraint.} We also constrain the estimation of surface normal by taking the ratio of $I(\phi_{pol})$ at polarizer angles $\phi_{pol} = 0 \hspace{0.1cm} \text{and} \hspace{0.1cm} \frac{\pi}{4}$ separately for diffuse and specular dominant regions. The following ratio constraint is inspired by the finding of Mecca \emph{et al.} \cite{mecca2017differential, logothetis2019differential}.
	\begin{equation}
	\centering
	\mathcal{L}_{ratio} = ||\alpha_{d}(F_{d}z_{x} - (-G_{d})z_{y})||_{1} + ||\alpha_{s}(F_{s}z_{y} - G_{s}z_{x})||_{1}
	\label{eq:13}
	\end{equation}
	Here, $F_{\star} = (-I(\frac{\pi}{4}) + A_{\star})$ and $G_{\star} = (I(0) - A_{\star} + \rho_{\star}A_{\star})$ with $\star \in \{d,s\}$, representing diffuse or specular case. Please refer to the supplementary material for detailed derivations of the constraints under diffuse and specular polarization dominance.\\
	The overall optimization objective is thus described as per Equation \ref{eq:14}.
	\begin{equation}
	\centering
	\mathcal{L}_{total} = \mathcal{L}_{rec} + \lambda_{geo}\mathcal{L}_{geo} + \lambda_{ratio}\mathcal{L}_{ratio}
	\label{eq:14}
	\end{equation}
	Here, $\lambda_{geo} = 1.0$ and $\lambda_{ratio} = 1.0$.
	
	\begin{figure}[t]
	\centering
	\includegraphics[width=\linewidth]{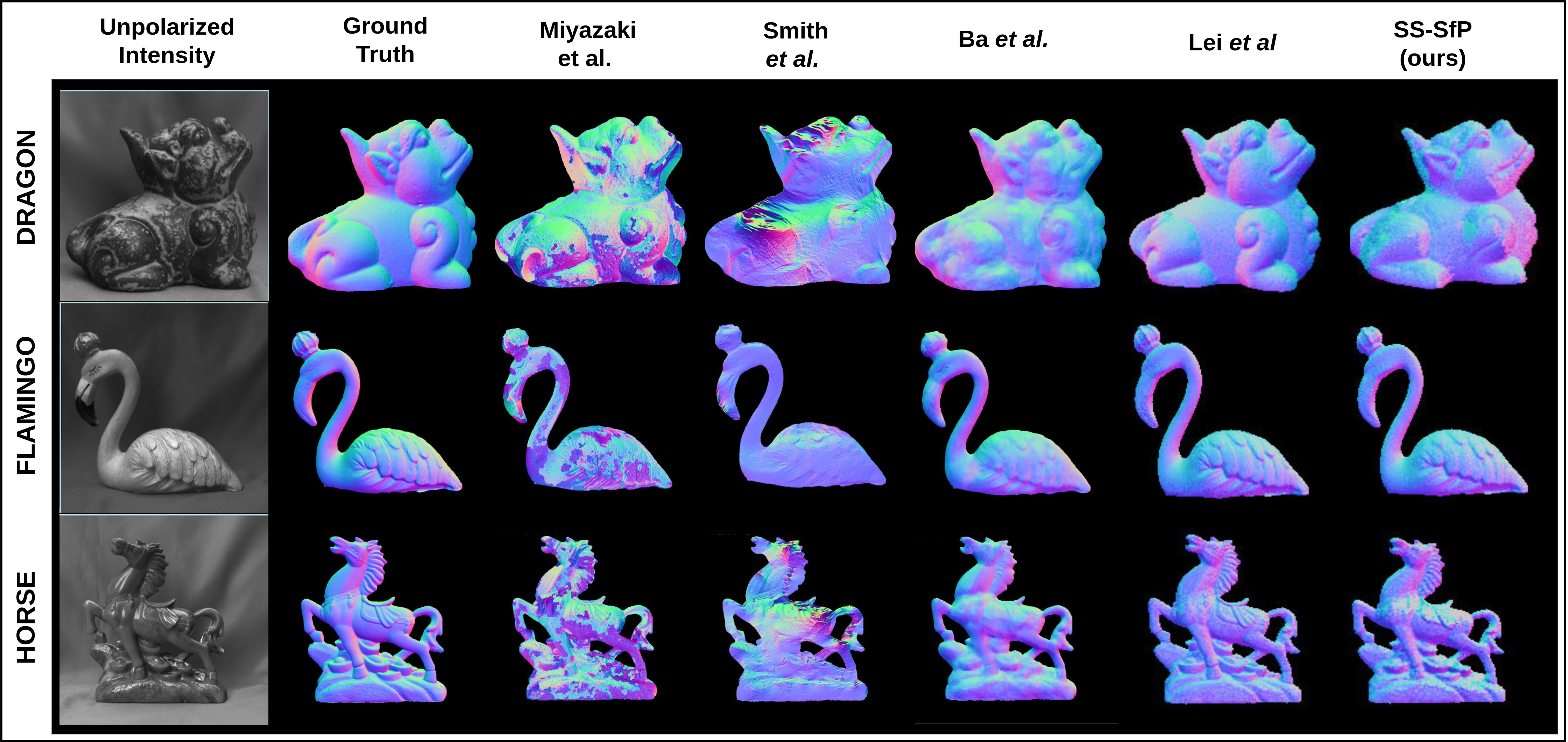}
	\caption{Qualitative results on surface normal estimation over a few objects from DeepSfP dataset \cite{ba2020deep}. }
	\label{fig:4}
	\end{figure}

	\section{Experimental Evaluation}
	\label{sec:res}
	In this section, we analyze the performance of the proposed framework and compare it against three physics-only methods \cite{smith2018height, miyazaki2003polarization, mahmoud2012direct} and two learning-based methods \cite{ba2020deep,lei2022shape} for SfP trained over two datasets, DeepSfP \cite{ba2020deep} and SPW \cite{lei2022shape}. The baseline methods have either assumed diffuse reflection \cite{smith2018height}, known lightings and/or albedos \cite{smith2018height, mahmoud2012direct}, or use ground truth normals for supervision \cite{ba2020deep, lei2022shape}. We report the widely used mean angular error (MAE) score (in degree) for quantitative analysis. Training of SS-SfP is not required as it learns at the test time through optimization.

	\subsection{Quantitative Evaluation} \label{sec:quant_eval}

	\begin{figure}[t]
		\centering
		\includegraphics[width=\linewidth]{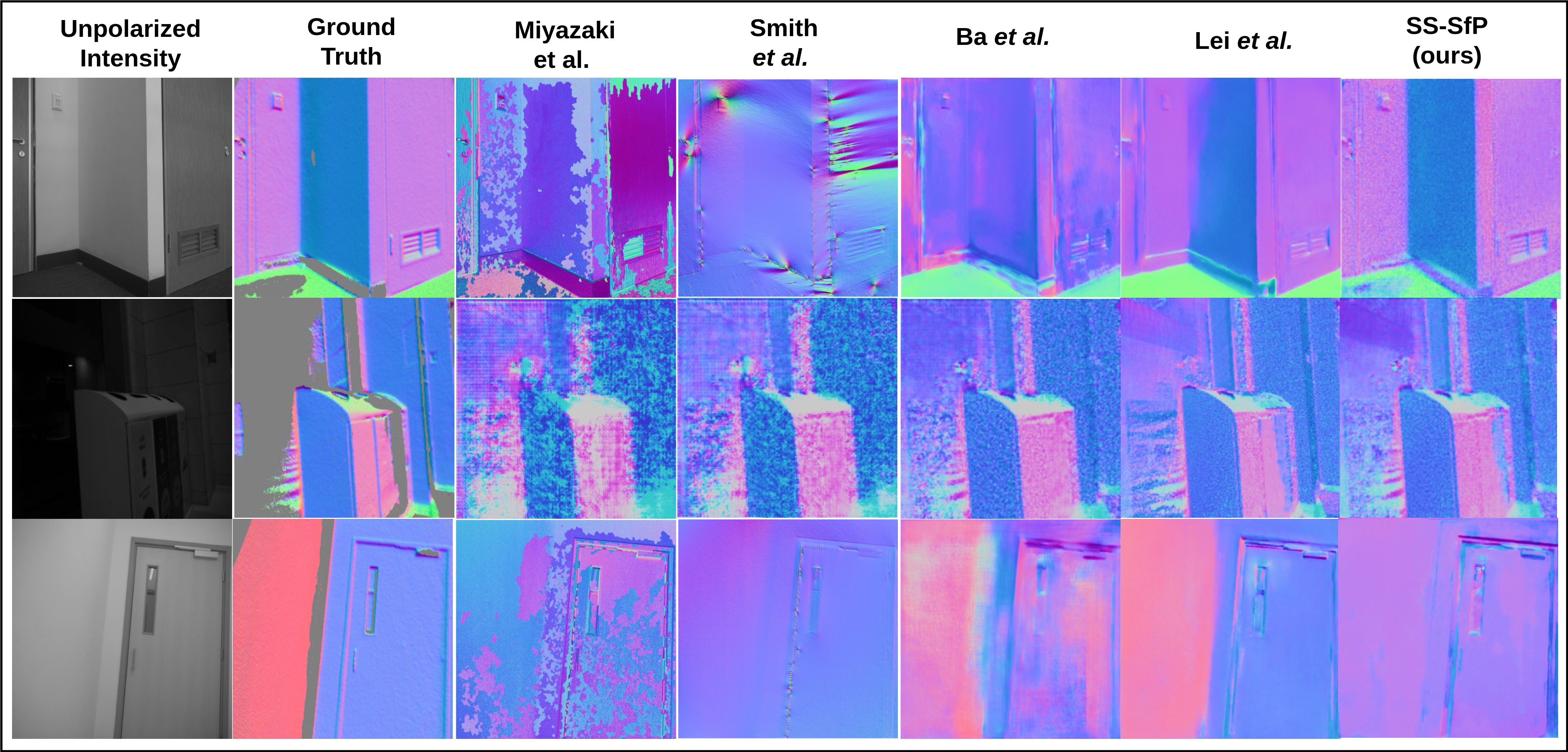}
		\caption{Qualitative results on surface normal estimation on a few scenes from SPW dataset \cite{lei2022shape}}
		\label{fig:5}
		\vspace{-0.75cm}
	\end{figure}
	
	\begin{table*}[t]
		\centering
		\resizebox{\textwidth}{!}{%
			\begin{tabular}{c|ccccc|cccc|c|c|c}\hline
				\multicolumn{1}{c}{\multirow{2}{*}{ID}} & \multicolumn{5}{|c|}{Encoder Input} & \multicolumn{4}{c|}{Decoder Input} & \multirow{2}{*}{\begin{tabular}[c]{@{}c@{}}Depth and Normal\\  \#Branches\end{tabular}} & \multicolumn{2}{c}{MAE} \\
				& Raw pol images & ($A$, $\rho$, $\phi$) & Normal Priors & ($A, \alpha_{d}, \alpha_{s}$) & VE & Encoder out & Normal Priors & ($A, \alpha_{d}, \alpha_{s}$) & VE &  & DeepSfP & SPW \\ \hline
				1 & \ding{51} & \ding{55} &  \ding{55} & \ding{55} & \ding{55} & \ding{51} & \ding{55}  & \ding{55} & \ding{55}  & 2 & 30.26 & 40.75   \\
				2 & \ding{51} & \ding{51} &  \ding{55} & \ding{55} & \ding{55} & \ding{51} & \ding{55}  & \ding{55} & \ding{55}  & 2 & 29.14 & 39.18\\
				3 & \ding{51} & \ding{55} &  \ding{51} & \ding{55} & \ding{55} & \ding{51} & \ding{55}  & \ding{55} & \ding{55}  & 2 & 20.98 & 32.75 \\
				4 & \ding{51} & \ding{55} &  \ding{55} & \ding{51} & \ding{55} &  
				\ding{51} &\ding{55}  & \ding{55} & \ding{55}  & 2 & 20.51 & 30.94 \\
				5 & \ding{51} & \ding{55} &  \ding{55} & \ding{51} & \ding{51} & \ding{51} & \ding{55}  & \ding{55} & \ding{55}  & 2 & 20.22 & 21.63 \\ 
				6 & \ding{51} & \ding{55} &  \ding{55} & \ding{55} & \ding{55} & \ding{51} & \ding{51}  & \ding{55} & \ding{51}  & 2 & 17.91 & 20.87\\
				7 & \ding{51} & \ding{55} &  \ding{55} & \ding{55} & \ding{55} & \ding{51} & \ding{55}  & \ding{51} & \ding{51}  & 2 & \textbf{16.89} & \textbf{19.77}\\
				8 & \ding{51} & \ding{55} &  \ding{55} & \ding{55} & \ding{55} & \ding{51} & \ding{55}  & \ding{51} (w/o SPADE) & \ding{51} (w/o SPADE) & 2 & 20.96 & 23.71 \\
				9 & \ding{51} & \ding{55} &  \ding{55} & \ding{55} & \ding{55} & \ding{51} & \ding{55}  & \ding{51} & \ding{51}  & 1 & 21.29 & 27.19 \\\hline
				10 & \multicolumn{10}{l|}{SS-SfP: Without instance normalization in the encoder} & 18.29 & 21.38\\
				11 & \multicolumn{10}{l|}{SS-SfP: Decoder with Self-Attention (as proposed in \cite{lei2022shape})} & 19.69 & 21.78\\
				12 & \multicolumn{10}{l|}{SS-SfP: without geometric constraint ($\mathcal{L}_{geo}$)} & 22.13 & 29.05\\
				13 & \multicolumn{10}{l|}{SS-SfP: without polarization angle ratio constraint ($\mathcal{L}_{ratio}$)} & 19.97 & 27.16\\\hline
				
			\end{tabular}%
		}
		
		\caption{Summary of ablation study over various design choices (ID 1-9) and architectural variations (ID 10-13) for the proposed framework.}
		\label{tab:4}
		\vspace{-0.5cm}
	\end{table*}
	\setlength{\tabcolsep}{1pt}
	
	\textbf{Evaluation Setting.} Our prime focus is to highlight the strength of the SS-SfP for self-supervised shape estimation. However, since there is no self-supervised framework available and for a fair comparison, we evaluate the learning-based baseline methods \cite{ba2020deep,lei2022shape} and the proposed SS-SfP under two settings: Supervised (S) and Self-supervised (SS) and compare them in the respective settings. Kindly refer to the supplementary material for details of training and evaluation under each setting.\\
	
		\begin{figure}[t]
		\centering
		\includegraphics[width=\linewidth]{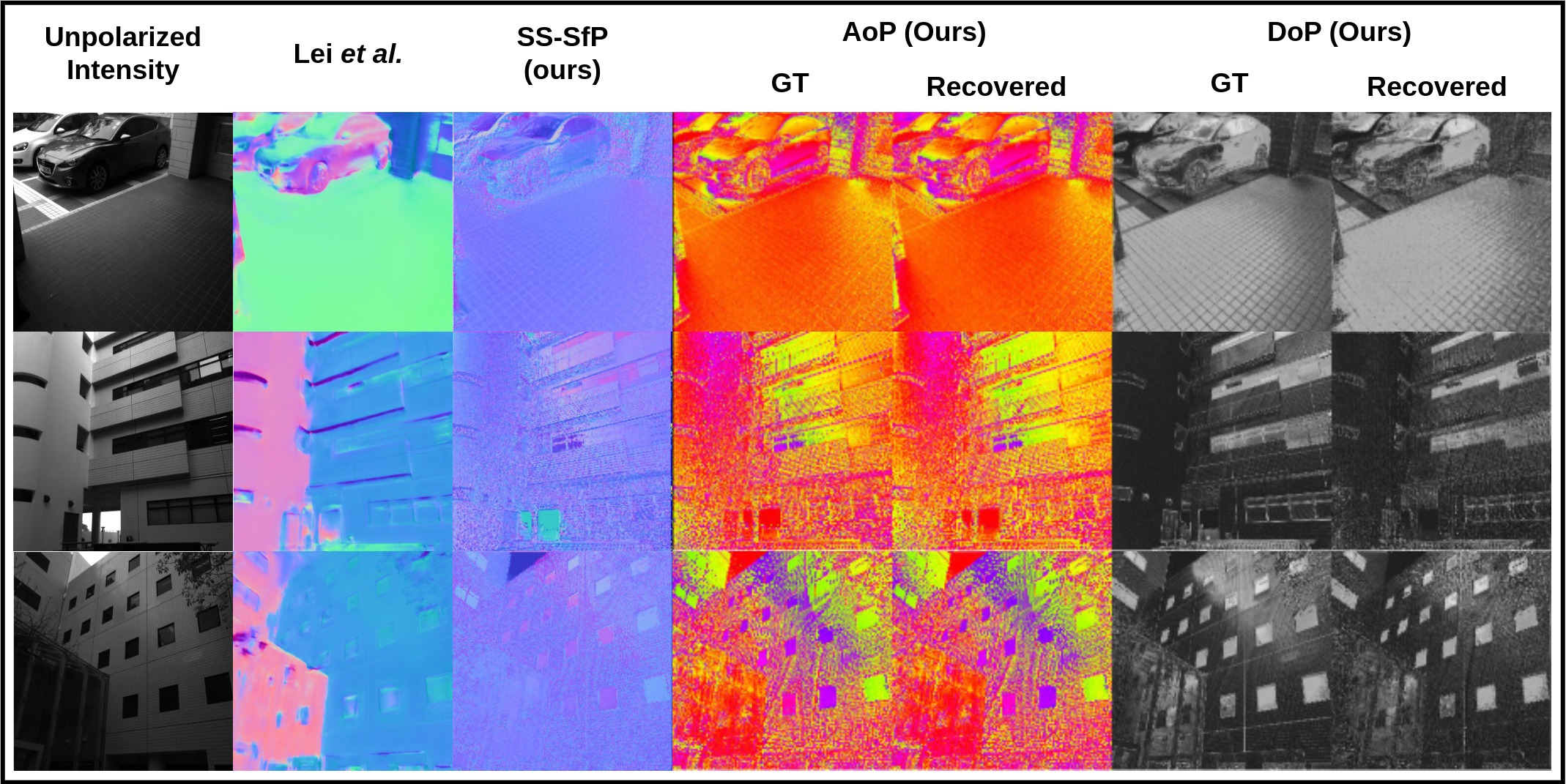}
		\caption{Qualitative results on far-field outdoor scenes. Since ground truth normals are unavailable, we validate the efficacy of estimated normals through the recovered AoP ($\widehat{\phi}$) and DoP ($\widehat{\rho}$).}
		\label{fig:6}
		\vspace{-0.8cm}
	\end{figure}

	\indent We find three key observations from Table \ref{tab:1}. (a) The proposed SS-SfP performs either the best or second best in either of the training settings. (b) SS-SfP outperforms all the baseline methods (across both the settings) except that of Lei \emph{et al.} \cite{lei2022shape} (under a fully supervised setting), where it falls short by a small margin owing to the benefit of direct supervision. Furthermore, while Lei \emph{et al.} \cite{lei2022shape} use the known camera intrinsics-based view encoding for handling non-orthographic projection (designed specifically for scene-level SfP), we employ a relatively simpler displacement field-based view-encoding. (c) The reduced performance of methods proposed by Ba \emph{et al.} \cite{ba2020deep} and Lei \emph{et al.} \cite{lei2022shape} under a self-supervised setting can be attributed to the fact that the network presumably finds an easier way to recover $\rho$ and $\phi$ directly from the input (remember, $\rho$ and $\phi$ are part of the input to \cite{ba2020deep, lei2022shape}) and focus less on the normal estimation since there is no direct supervision. Figure \ref{fig:6} shows a qualitative comparison of the method by Lei \emph{et al.} \cite{lei2022shape} and SS-SfP under self-supervised setting since the ground truth normals are not present for those far-field outdoor scenes.\\\indent
	Table \ref{tab:2} examines the performance of each object in the DeepSfP dataset. Table \ref{tab:3} shows the robustness of the proposed method to different lighting conditions (also, Figure \ref{fig:6}). We find that SS-SfP exhibits only slight variation in MAE across indoor and outdoor overcast lighting. However, it shows a little higher MAE over outdoor solar lighting. This is primarily because the observed diffuse reflection dominates over the specular ones due to the high intensity of incoming light from the sun, leading to a slightly poor judgement about the reflection-dependent normal orientation. Furthermore, while we assume unpolarized incident illumination, the illumination incident on the scene under outdoor lighting could be partially polarized due to the reflections from other surfaces (acting as potential light sources). Overall, the framework yields an \underline{average SSIM} of $\mathbf{0.836}$ and $\mathbf{0.714}$  on the recovered DoP, an \underline{average SSIM} of $\mathbf{0.917}$ and $\mathbf{0.822}$ on the recovered polarization images, and a \underline{mean absolute error} of $\mathbf{0.961^{\circ}}$ and $\mathbf{1.772^{\circ}}$ on the recovered AoP over the DeepSfP \cite{ba2020deep} and SPW \cite{lei2022shape} test sets, respectively.

	\subsection{Qualitative Evaluation}
	Figure \ref{fig:4} and \ref{fig:5} show the qualitative comparison of SS-SfP over the other baseline methods. In Figure \ref{fig:4}, we show the qualitative results on three complex objects: DRAGON, FLAMINGO, and HORSE in the DeepSfP dataset. We observe that SS-SfP is not too far from a completely supervised method by Lei \emph{et al.} and performs better for some scenes (row 1, Figure \ref{fig:5}) and nearly equal for objects in rows 2 and 3, Figure \ref{fig:4}. SS-SfP can also scale to far-field outdoor scenes (see Figure \ref{fig:6}) with distances far beyond the depth captured in the SPW dataset, leveraging the fact that the relationship between polarized light and surface normals remains unaffected by distance. While the normal estimates exhibit limited directional variations despite better AoP and DoP reconstructions (one of the limitations), they capture the finer geometrical variations in the scenes, such as tile partitions on the floor, glass shields on the car, and glass windows on the building (Figure \ref{fig:6}), which are not prominent in the results obtained by Lei \emph{et al.} \cite{lei2022shape}. This indicates the ability of the network to learn the shape from polarization images, especially when ground truth capture is difficult and inaccurate, and validates that SfP does not deviate from mere image semantics. Moreover, we observe that extreme high-frequency variations in the shapes are not captured well compared to their supervised counterparts (another limitation), leading to higher MAE in Table \ref{tab:1} and \ref{tab:2} and some blocky artefacts, especially in HORSE and DRAGON of the DeepSfP dataset (Figure \ref{fig:4}). This behavior is generally observed for optimization-based self-supervised approaches.
	
	\section{Ablation and Discussion} 
	\label{sec:ablation}
	
	Table \ref{tab:4} summarizes the effect of several design choices and potential variants of the proposed framework.\\
	\textbf{Effect of reflectance cues and view encoding.} We obtain lower MAE when reflectance cues and view encoding are injected into the decoder than when they are provided as input to the encoder (compare IDs $5$ and $7$, Table \ref{tab:4}). Since such deep residual architectures tend to wash away necessary information obtained from the input \cite{huang2016deep, ba2020deep}, it becomes essential to introduce (or re-introduce) guiding information to the decoder. View encoding subsumes the impact of viewing direction on the polarization data (note the difference of $9.31^{\circ}$ among IDs $4$ and $5$).\\
	\textbf{Effect of SPADE normalization.} We use SPADE normalization for two reasons: (a) it is a generalization of several other existing normalization schemes \cite{park2019semantic} and (b) it is suited for better image understanding and synthesis through semantic layouts (diffuse and specular separation, in our case). We find that SPADE normalization helps the decoder better manage the reflectance cues and handle ambiguities for surface normal estimation through adaptive modulation parameters. In general, we observe better performance with SPADE normalization, irrespective of the information fed at its input (compare IDs $5, 6, 8$ with $7$, Table \ref{tab:4}).\\
	\textbf{Physics-based normal priors vs reflectance cues.} We find that the performance with reflectance cues is better than that when using normal priors (Table \ref{tab:4} (ID $6$ and $7$)) although the difference in MAE is just around $1^{\circ}$. Moreover, we find additional concerns with such handcrafted priors on normals. (a) Normal priors derived using reflectance-specific DoP \cite{collett2005field} inherently have ambiguous angles since no such model characterizes the priors on mixed reflections. (b) They are sensitive to noise present in the raw polarization images. Merely shifting the azimuth angles by $\pi$ or $\pi/2$ will not recover proper surface normals from noisy images. (c) Their computation is time-consuming and leads to slower inference (see Table \ref{tab:2}, last column).\\
	\textbf{Effect of estimating depth.} The depth is estimated to devise a geometrical constraint (used in literature) for better normal estimates (ID $7$, Table \ref{tab:4}). However, estimating depth directly shows poor performance (see ID $9$, Table \ref{tab:4}) and is mainly attributed to the discontinuities offered by the differentiation step in the depth estimates (and thus, the surface normal map) \cite{yu2017shape}.\\
	Additional ablations studies, implementation details, qualitative results, and other necessary information are detailed in the supplementary material.

	\vspace{-0.25cm}
	\section{Conclusion}
	\label{sec:conc}
	We proposed SS-SfP - a neural inverse rendering-based self-supervised framework to address SfP under mixed reflections by obtaining per-pixel diffuse and specular reflectance components and refractive index estimation in a completely self-supervised manner. We evaluated SS-SfP under both supervised and self-supervised settings with a heavy emphasis on self-supervised learning. Further, SS-SfP is also shown to extend to far-field outdoor scenes and opens up a direction toward in-the-wild geometry reconstruction from polarization images. We believe this work would generate more traction towards self-supervised learning methods to address SfP under mixed reflections without needing 3D ground truth.\\
	\noindent\textbf{Acknowldegement.} This work is supported by the Prime Minister Research Fellowship (PMRF) awarded to Ashish Tiwari and Jibaben Patel Chair in AI awarded to Shanmuganathan Raman.

\newpage

\section*{Supplementary}


	\noindent The following is the summary of the contents covered in the supplementary material.
	\begin{enumerate}
		\item Ambiguities in SfP
		\item Implementation details
		\item Dataset Details
		\item Additional Ablation Experiments
		\item Additional Qualitative Results
		\item Polarization Angle Ratio Constraint - Derivation
	\end{enumerate}

  \begin{figure}[h]
     \includegraphics[width=\linewidth]{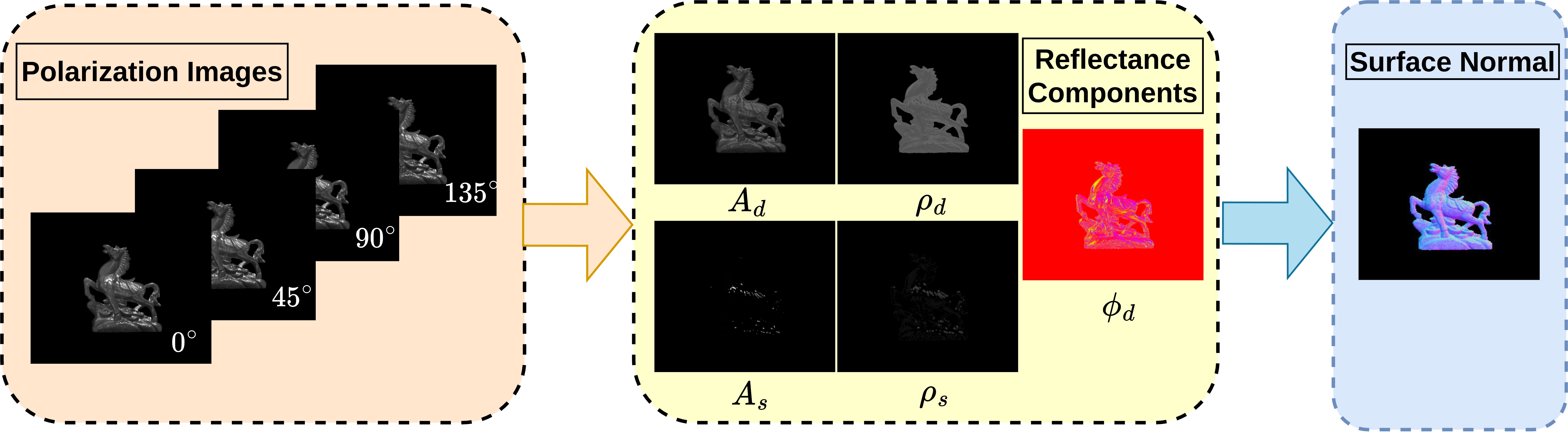}
		\centering
		\caption{Overview of the proposed self-supervised inverse rendering-based framework (SS-SfP) to obtain per-pixel surface normals under mixed polarization by decomposing the diffuse ($A_{d}$) and specular ($A_{s}$) reflection components from the raw polarization images.}
		\label{fig:teaser}
 \end{figure}
	
	\begin{figure*}[t]
		\centering
		\includegraphics[width=\linewidth]{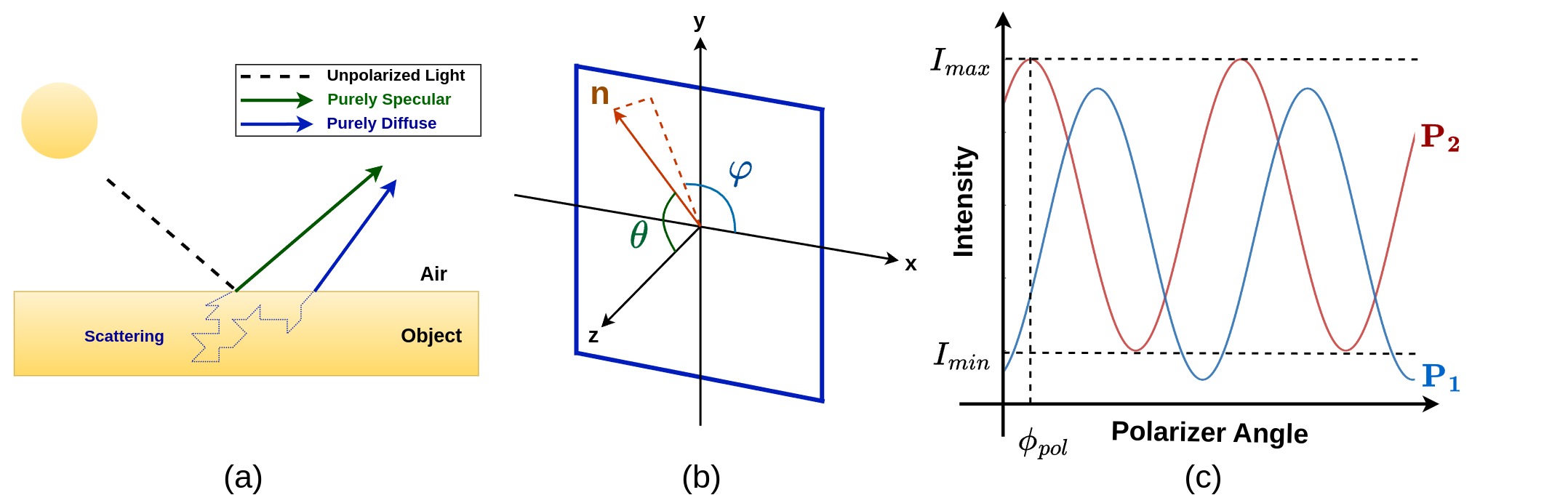}
		\caption{(a) Mixed reflections (and polarization) off the surface. (b) Coordinate system for polarization imaging. (c) Transmitted Radiance Sinusoid (TRS) showing the observed intensities under varying polarizer angles for two pixels ($P_{1}$ and $P_{2}$) with different surface normals.}
		\label{fig:prelims}
	\end{figure*}
	
	\section*{1. Ambiguities in SfP} \label{sec:amb}
	An unpolarized light striking a surface point exhibits diffuse and/or specular (mixed) reflection (see Figure \ref{fig:prelims} (a)). The estimation of $\phi$ and $\rho$ depends on the surface reflectance model and directly relates to the azimuth $(\varphi)$ and the zenith $(\theta)$ angles, respectively, described as per the coordinate system shown in Figure \ref{fig:prelims}(b). The polarization image formation is generally given by Equation \ref{eq:main1}.
	\begin{equation}
	\centering
	I(\phi_{pol}) = \text{A} + \text{B} \cos(2\phi_{pol}-2\phi)
	\label{eq:main1}
	\end{equation}
	Equation \ref{eq:main1} is manifested in the form of a Transmitted Radiance Sinusoid (TRS), as shown in Figure \ref{fig:prelims} (c). 
	
	\textbf{(i) Azimuth Angle Ambiguities.}
		As per Equation \ref{eq:main1}, two azimuth angles separated by $\pi$ radians cannot be distinguished in polarization images, i.e., $\varphi$ and $\varphi + \pi$ will have the same result. This is referred to as \textit{\textbf{azimuthal angle ambiguity}}. Consider $T$ and $R$ as the transmittance and reflectance coefficients either parallel $(||)$ or perpendicular $(\perp)$ to the incidence plane. Under diffuse reflection, a portion of the light enters the object and gets refracted and thus, partially polarized \cite{wolff1993constraining} with a greater magnitude in the direction parallel to the incidence plane $(T_{||} > T_{\perp})$. Therefore, maximum light intensity is observed for $\varphi = \phi$. Under specular reflection, the reflected light is predominantly polarized in the direction perpendicular to the incidence plane $(R_{\perp} > R_{||})$. Therefore, the maximum light intensity will be observed at $\varphi = \phi \pm \frac{\pi}{2}$. In short, for a general surface, when the type of reflectance is not known apriori, we are unsure if the estimated angle should be shifted $\pi/2$. This is called \textit{\textbf{azimuthal model mismatch}}.
	
	\textbf{(ii) Zenith Angle Ambiguities.}
		The zenith angle relies on the degree of polarization (DoP) $(\rho)$ and refractive index $(\eta)$. Moreover, as in the case of azimuthal angle estimation, the type of reflectance model affects the zenith angle estimation as well and produces \textit{\textbf{zenith model mismatch}}, as described below. 
		The DoP is described as per Equation \ref{eq:3} for diffuse reflection \cite{collett2005field}.
	\begin{equation}
	\centering
	\rho = \frac{(\eta - \frac{1}{\eta})^{2}\sin^{2}\theta}{2 + 2\eta^{2} - (\eta + \frac{1}{\eta})^{2}\sin^{2}\theta + 4\cos\theta \sqrt{\eta^{2}-\sin^{2}\theta}}
	\label{eq:3}
	\end{equation}
	Similarly, the DoP is described in Equation \ref{eq:4} for specular reflection \cite{collett2005field}.
	\begin{equation}
	\centering
	\rho = \frac{2\sin\theta \tan\theta \sqrt{\eta^{2}-\sin^{2}\theta}}{\eta^{2} - 2\sin^{2}\theta + \tan^{2}\theta}
	\label{eq:4}
	\end{equation}
	However, this relation applies to highly specular objects and has been used for metallic objects. The very requirement of known refractive index $(\eta)$ imposes \textit{\textbf{refractive distortion}} if an improper refractive index is used. Moreover, for regions having a zenith angle close to zero, DoP is small, and estimated surface normals are noisy due to low SNR. The readers are requested to kindly refer to \cite{shi2020recent} for more details.
	
	\section*{2. Implementation Details}
	The network weights are randomly initialized just at the beginning, and the weights are subsequently updated through the loss functions for $2500$ iterations. The framework is trained over $256 \times 256$ sized images and is implemented in PyTorch \cite{paszke2017automatic} over the NVIDIA RTX $5000$ GPU with $16$ GB memory. Each object or scene is optimized using Adam optimizer \cite{kingma2014adam} with default parameters. Note that the initial learning rate for the optimization is set to $0.001$. The network is optimized for $2500$ iterations with a learning rate decay of $0.1$ after every $250$ iterations.
	
	To train the baselines under a self-supervised setting, we use their respective networks for normal estimation and recover the polarization information $(\rho, \phi)$ from the estimated normals. We replace the normal supervision with the reconstruction error between estimated and ground truth AoP and DoP. We could not enforce the geometry and ratio constraints since the two methods \cite{ba2020deep, lei2022shape} do not estimate depth. Further, under a supervised setting, SS-SfP is trained over the respective train sets under direct normal supervision and tested over the respective test sets. We stick to the same train-test split as originally given for the respective datasets for fair comparison so that they do not contain images from the same scene.

	\section*{3. Dataset Details}
	
	\hspace{4mm}\textbf{DeepSfP Dataset}\cite{ba2020deep} contains $33$ objects in total out of which $25$ objects are kept for the training while the remaining $8$ belong to the test set. Each of the objects are imaged under $3$ different lighting conditions (indoor, outdoor-sunny day, and outdoor-cloudy day) and $4$ different orientations (front, back, left, and right) such that we have a total of $300$ images in the train set.
	
	\textbf{SPW Dataset}\cite{lei2022shape} contains the scene-level polarization data for the scenes in the wild. It consists of $522$ images from $110$ different scenes with diverse object materials and lighting conditions. It contains $403$ images in the train set and $119$ in the test set.

	\section*{4. Additional Ablation Experiments}
	Table \ref{tab:5} reports the variation in the performance of the proposed framework with the number of layers in the encoder and the decoder. The network performance is best for $6$ and $5$ blocks (each for encoder and decoder) over the DeepSfP and SPW datasets, respectively. However, we finally resorted to $5$ blocks for a lighter network. 
	
	\begin{table*}[t]
		\centering
		\resizebox{\textwidth}{!}{%
			\begin{tabular}{c|ccccc|cccc|c|c|c}\hline
				\multicolumn{1}{c}{\multirow{2}{*}{ID}} & \multicolumn{5}{|c|}{Encoder Input} & \multicolumn{4}{c|}{Decoder Input} & \multirow{2}{*}{\begin{tabular}[c]{@{}c@{}}Depth and Normal\\  \#Branches\end{tabular}} & \multicolumn{2}{c}{MAE} \\
				& Raw pol images & ($A$, $\rho$, $\phi$) & Normal Priors & ($A, \alpha_{d}, \alpha_{s}$) & VE & Encoder out & Normal Priors & ($A, \alpha_{d}, \alpha_{s}$) & VE &  & DeepSfP & SPW \\ \hline
				1 & \ding{51} & \ding{55} &  \ding{55} & \ding{55} & \ding{55} & \ding{51} & \ding{55}  & \ding{55} & \ding{55}  & 2 & 30.26 & 40.75   \\
				2 & \ding{51} & \ding{51} &  \ding{55} & \ding{55} & \ding{55} & \ding{51} & \ding{55}  & \ding{55} & \ding{55}  & 2 & 29.14 & 39.18\\
				3 & \ding{51} & \ding{55} &  \ding{51} & \ding{55} & \ding{55} & \ding{51} & \ding{55}  & \ding{55} & \ding{55}  & 2 & 20.98 & 32.75 \\
				4 & \ding{51} & \ding{55} &  \ding{55} & \ding{51} & \ding{55} &  
				\ding{51} &\ding{55}  & \ding{55} & \ding{55}  & 2 & 20.51 & 30.94 \\
				5 & \ding{51} & \ding{55} &  \ding{55} & \ding{51} & \ding{51} & \ding{51} & \ding{55}  & \ding{55} & \ding{55}  & 2 & 20.22 & 21.63 \\ 
				6 & \ding{51} & \ding{55} &  \ding{55} & \ding{55} & \ding{55} & \ding{51} & \ding{51}  & \ding{55} & \ding{51}  & 2 & 17.91 & 20.87\\
				7 & \ding{51} & \ding{55} &  \ding{55} & \ding{55} & \ding{55} & \ding{51} & \ding{55}  & \ding{51} & \ding{51}  & 2 & \textbf{16.89} & \textbf{19.77}\\
				8 & \ding{51} & \ding{55} &  \ding{55} & \ding{55} & \ding{55} & \ding{51} & \ding{55}  & \ding{51} (w/o SPADE) & \ding{51} (w/o SPADE) & 2 & 20.96 & 23.71 \\
				9 & \ding{51} & \ding{55} &  \ding{55} & \ding{55} & \ding{55} & \ding{51} & \ding{55}  & \ding{51} & \ding{51}  & 1 & 21.29 & 27.19 \\\hline
				10 & \multicolumn{10}{l|}{SS-SfP: Without instance normalization in the encoder} & 18.29 & 21.38\\
				11 & \multicolumn{10}{l|}{SS-SfP: Decoder with Self-Attention (as proposed in \cite{lei2022shape})} & 19.69 & 21.78\\
				12 & \multicolumn{10}{l|}{SS-SfP: without geometric constraint ($\mathcal{L}_{geo}$)} & 22.13 & 29.05\\
				13 & \multicolumn{10}{l|}{SS-SfP: without polarization angle ratio constraint ($\mathcal{L}_{ratio}$)} & 19.97 & 27.16\\\hline
				
			\end{tabular}%
		}
		
		\caption{Summary of quantitative ablation study over various design choices (ID 1-9) and architectural variations (ID 10-13) for the proposed framework (repeated from main paper).}
		\label{tab:abl}
	\end{table*}
	\setlength{\tabcolsep}{1pt}
	
	\begin{table}[h]
		\centering
		\resizebox{0.8\linewidth}{!}{%
			\begin{tabular}{c|cc} \hline
				\multirow{2}{*}{\begin{tabular}[c]{@{}c@{}}\#Encoder and Decoder \\ Blocks\end{tabular}} & \multicolumn{2}{c}{\begin{tabular}[c]{@{}c@{}}MAE (in deg.)\end{tabular}} \\ 
				& DeepSfP & SPW \\\hline
				2 & 28.57 & 33.07 \\
				3 & 19.18 & 26.35 \\
				4 & 18.02 & 21.64 \\
				5 & 16.89 & \textbf{19.77} \\
				6 & \textbf{16.84} & 20.14 \\
				8 & 18.23 & 21.78 \\
				10 & 19.11 & 22.16 \\\hline
			\end{tabular}%
		}
		\caption{Ablation experiments for the number of encoder and decoder blocks on the DeepSfP and the SPW datasets. We choose $5$ blocks each for the encoder and the decoder in our model
			according to these quantitative results.}
		\label{tab:5}
		\vspace{-0.5cm}
	\end{table}
	
\begin{figure}[h]
		\centering
		\includegraphics[width = 0.95\linewidth]{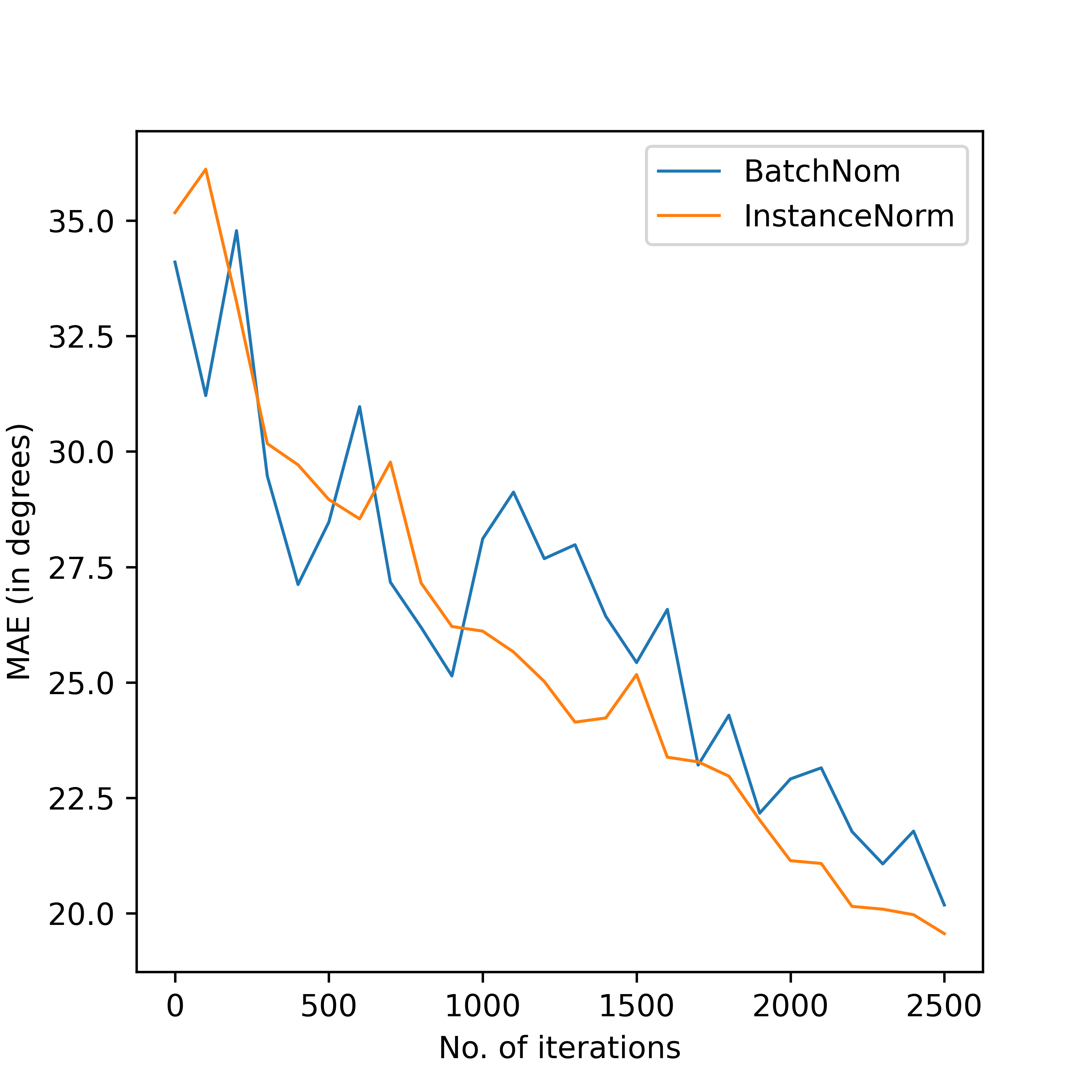}
		\caption{Learning curve of the proposed framework over the SPW dataset \cite{lei2022shape}}
		\label{fig:8}
		\vspace{-0.2cm}
	\end{figure}

	Further, we chose to use instance normalization after observing a relatively smoother and faster convergence with instance normalization when compared to that with batch normalization, as shown in Figure \ref{fig:8}. The values are averaged over the scenes in the test set of the SPW dataset \cite{lei2022shape}. Table \ref{tab:abl} shows how instance normalization achieves better performance. Observing a slight fallback in performance compared to SPW \cite{lei2022shape}), we tried an interesting variant to use self-attention \cite{chen2021transunet, yang2021transformer} (see ID 11, Table \ref{tab:abl}) in the decoder (inspired by SPW \cite{lei2022shape}). However, the performance still suffered when compared to the proposed SS-SfP.
	
	\begin{figure}[h]
		\centering
		\includegraphics[width = \linewidth]{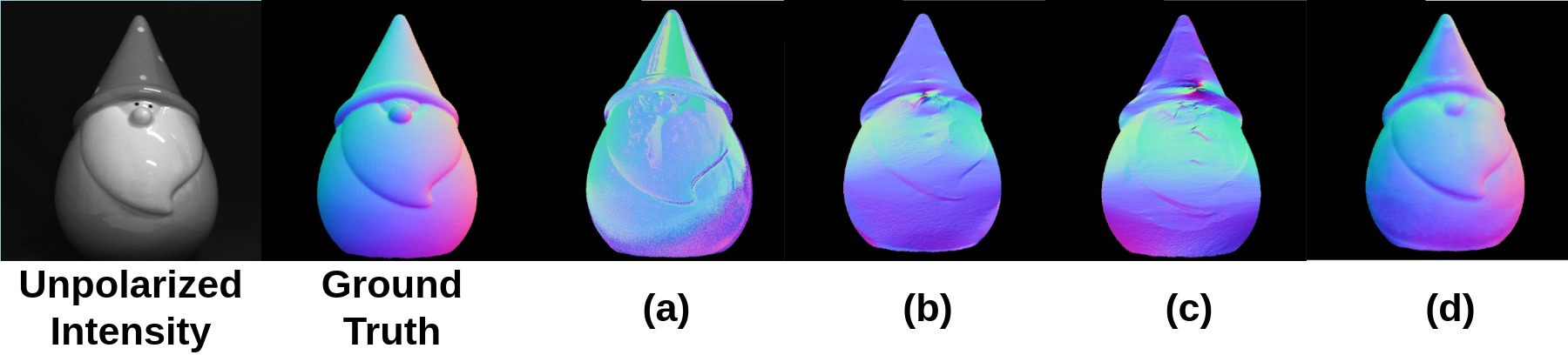}
		\caption{Qualitative effect of specific design choices on the network performance. (a) without injecting the reflectance cues into the decoder. (b) without reflectance cues, with $\mathcal{L}_{ratio}$. (c) with total variation loss (d) with reflectance cues and $\mathcal{L}_{ratio}$ (ours). Note that $\mathcal{L}_{geo}$ is included in experiments (a), (b), (c), and (d).  }
		\label{fig:9}
		\vspace{-0.5cm}
	\end{figure}

	Figure \ref{fig:9} shows the quality of surface normals estimates under different design choices. We find that the high-frequency details are blurred out if we do not inject the reflectance cues (Figure \ref{fig:9} (a)). Further, since the scenes are mostly diffuse-dominant, the network fails to estimate normals in the specular regions precisely. While the polarization angle ratio constraint does seem to help a bit (without reflectance cues), it also fails at the highly specular regions (Figure \ref{fig:9} (b)). Moreover, adding total variation loss smoothens out the surface normals (Figure \ref{fig:9} (c)). Therefore, we chose to inject the reflectance cues and used the geometric and ratio constraint for better surface normal estimates (Figure \ref{fig:9} (d)).
	
	\textbf{Why should we reconstruct $\phi$ and $\rho$?} The simple reason is to allow the network to model a perfect relation between surface normals and the physically measurable quantities: DoP ($\rho$) and AoP $(\phi)$ under mixed polarization and handle the underlying ambiguities. Furthermore, the derived quantities - $\phi_{d}, \phi_{s}, \rho_{d}, \text{and}, \rho_{s}$, can only be measured if the surface is purely diffuse or specular, which is seldom the case, and that too with the $\pi$-ambiguity. While there are closed-form expressions to establish such a relation for diffuse and specular reflections individually (see Section \ref{sec:amb}), there is no such model concerning mixed polarization. By reconstructing $\phi$ and $\rho$ from the surface normal estimates, we force the network to learn their inter-dependencies and further use them to reconstruct the polarization images as per the standard polarization image formation model, as described by Equation \ref{eq:main1}.

	\textbf{Why do we estimate both surface normals and depth?} Since surface normals can be obtained from depth derivatives, we could have a single decoder in a deep network for surface normal estimates. However, as discussed in the main paper, we observe poor performance quantitatively through just depth estimation (see ID $8$, Table \ref{tab:abl}). This is attributed to the discontinuities offered by the differentiation step in the depth estimates (and thus, surface normal map) \cite{yu2017shape}. One way could be to use smoothness constraints such as minimizing total variation. However, they were found to flatten out the normals (and smoothen out the high-frequency details), especially when there is no direct supervision for surface normals. Moreover, such self-supervised frameworks get unstable when applied to real data (such as spikes in the depth maps) \cite{yu2017shape}. Therefore, we estimate surface normal and depth through two different branches and enforce geometric constraint $(\mathcal{L}_{geo})$ and reflection-dependent ratio constraint $(\mathcal{L}_{ratio})$ for better results (see IDs $12$ and $13$, Table \ref{tab:abl}).

	\section*{5. Additional Qualitative Results}
\begin{figure}[t]
		\centering
		\includegraphics[width = \linewidth]{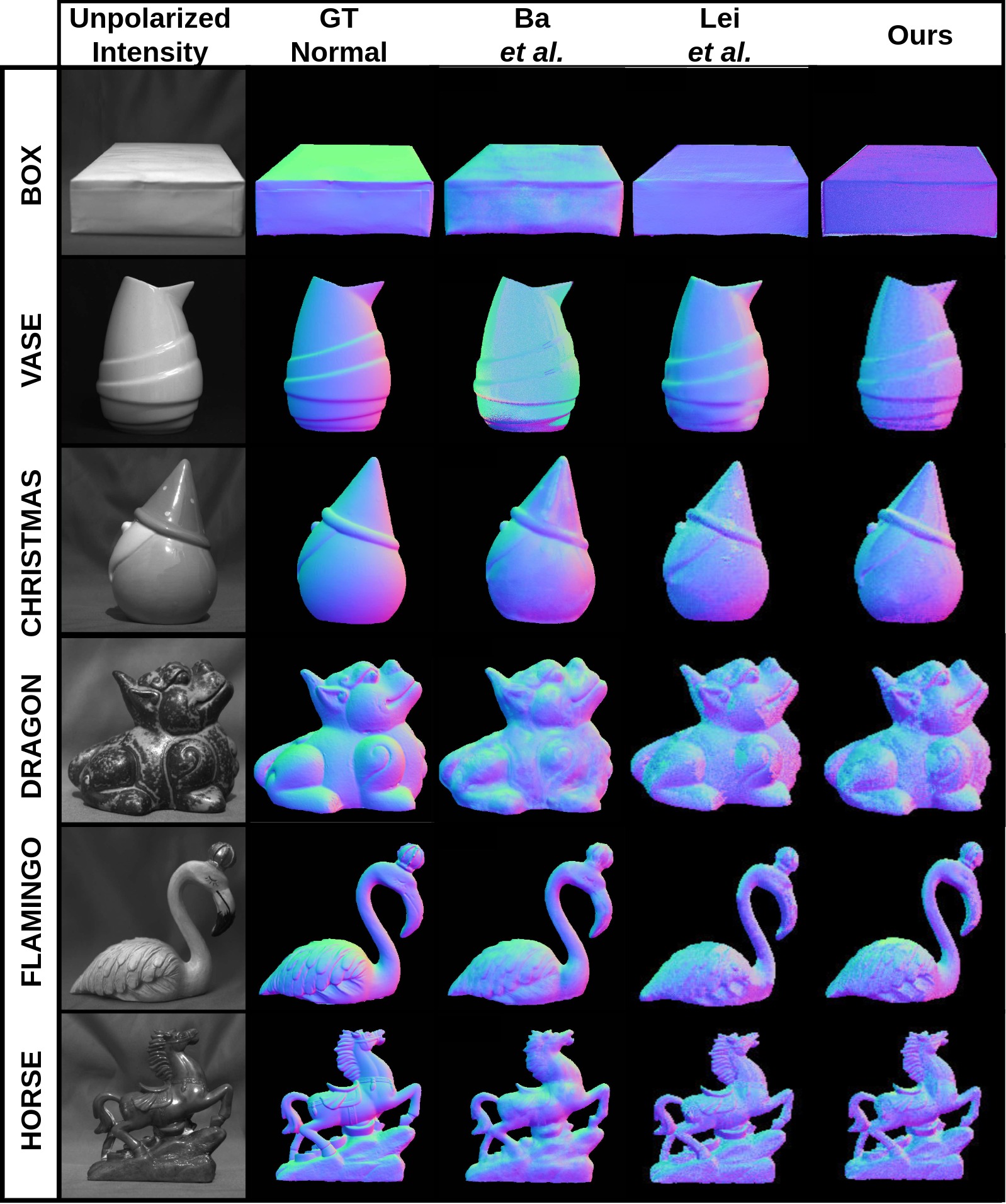}
		\caption{Additional qualitative results on the test set of the DeepSfP dataset \cite{ba2020deep}. }
		\label{fig:10}
		\vspace{-0.5cm}
	\end{figure}

	Figure \ref{fig:10} shows the qualitative results on three additional objects (BOX, VASE, and CHRISTMAS) of the DeepSfP dataset \cite{ba2020deep} that were not included in the main paper. Further, it also shows the results on the other three objects (DRAGON, FLAMINGO, and HORSE) observed from different views.
	
\begin{figure*}[h]
		\centering
		\includegraphics[width = \textwidth]{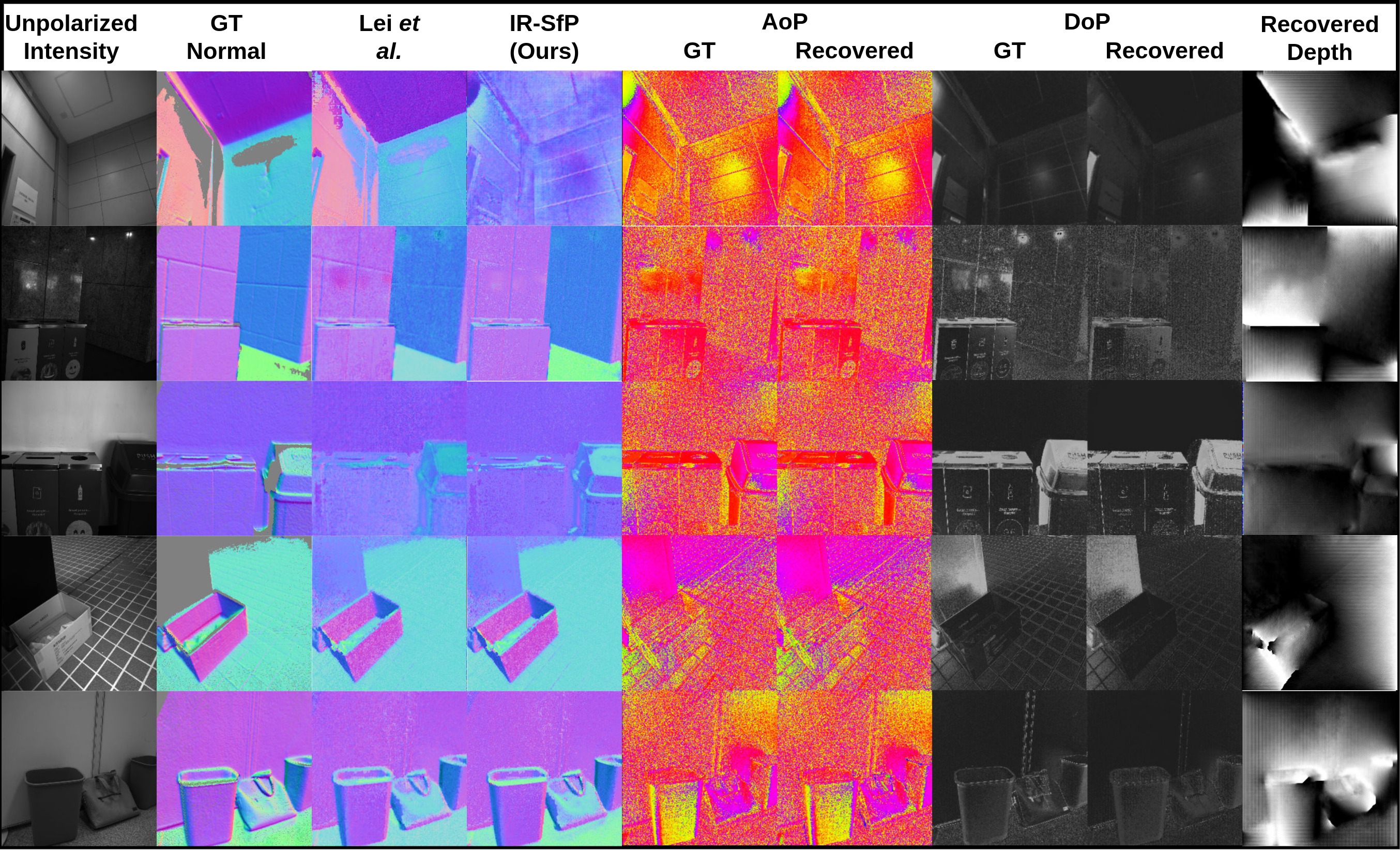}
		\caption{Additional qualitative results on the test set of the SPW dataset \cite{lei2022shape}. We also demonstrate the recovered phase angle (AoP), degree of polarization (DoP), and coarse depth maps. }
		\label{fig:11}
		
	\end{figure*}

	Figure \ref{fig:11} shows the results on some additional scenes chosen from the test set of the SPW dataset \cite{lei2022shape}. The proposed framework performs better than that of Lei \emph{et al.} for scenes in rows $1$ and $2$) and almost equally well for the scenes in rows $4$ and $5$ of Figure \ref{fig:11}. To validate the performance under a self-supervised setting, we also show the associated phase angle and degree of polarization in Figure \ref{fig:11}.
	
	\section*{6. Polarization Angle Ratio Constraint - Derivation}
	Let us start with the image formation model described in the main paper in Section 4.2 (Equation 7). 
	\begin{equation}
	\centering
	I(\phi_{pol}) = A_{m} + B_{m}cos(2\phi_{pol}-2\phi)
	\label{eq:16}
	\end{equation}
	We deploy the findings of \cite{logothetis2019differential,mecca2017differential} through a differential formulation of SfP to model the polarization angle ratio constraint.
	
	\subsection*{6.1 Diffuse Polarization}
	Let us consider the image formation for diffuse polarisation and expand the cosine term to get the following.
	\begin{align}
	\centering
	I(\phi_{pol}) = & \hspace{0.2cm} A_{d} + B_{d}\bigg(cos(2\phi_{pol})\Big(2cos^{2}(\phi_{d})-1\Big) \notag \\ & \hspace{0.2cm} + 2sin(2\phi_{pol})sin(\phi_{d})cos(\phi_{d})\bigg)
	\end{align}
	
	The first two components of the non-unit normal vector to the surface $\widehat{\mathbf{n}} = (n_{x}, n_{y}, n_{z})$ are proportional to $\nabla z$ up to a factor depending on the focal length $f$ such that $\mathbf{n} = \frac{\widehat{\mathbf{n}}}{||\widehat{\mathbf{n}}||} =\begin{bmatrix} g(f)z_{x} & g(f)z_{y} & -1 \end{bmatrix}^{T} = \begin{bmatrix} \text{sin}(\theta) \text{cos}(\phi)  & \text{sin}(\theta) \text{sin}(\phi) & \text{cos}(\theta)\end{bmatrix}^{T}$. By substituting for $cos(\phi)$ and $sin(\phi)$ at $\phi=\phi_{d}$, we obtain the following.
	\begin{align}
	\centering
	I(\phi_{pol}) = & \hspace{0.2cm} A_{d} + B_{d}\bigg(cos(2\phi_{pol})\Big(2\frac{z_{x}^{2}}{||\widehat{\mathbf{n}}||^{2}sin^{2}(\theta)}-1\Big) \notag \\ & \hspace{0.2cm} + 2sin(2\phi_{pol})\frac{z_{x}z_{y}}{||\widehat{\mathbf{n}}^{2}sin^{2}(\theta)}\bigg)
	\label{eq:18}
	\end{align}
	Here, for ease of understanding, we consider $g(f)=1$ (orthographic case). However, the same set of constraints also applies to the perspective case since the factor gets canceled out while taking the ratio. Simplifying the Equation \ref{eq:18}, we get,
	\begin{align}
	\centering
	I(\phi_{pol}) \hspace{0.1cm} - &\hspace{0.1cm}A_{d} + B_{d}cos(2\phi_{pol}) = \notag \\ & B_{d}\bigg(cos(2\phi_{pol})z_{x} + sin(2\phi_{pol})z_{y}\bigg)\frac{2z_{x}}{||\widehat{\mathbf{n}}||^{2}sin^{2}(\theta)}
	\label{eq:19}
	\end{align}
	Now, we consider the ratio of the Equation \ref{eq:19} evaluated at two polarizer angles $\phi^{1}_{pol}$ and $\phi^{2}_{pol}$.
	\begin{align}
	\centering
	\frac{I(\phi^{1}_{pol}) - A_{d} + B_{d}cos(2\phi^{1}_{pol})} {I(\phi^{2}_{pol}) - A_{d} + B_{d}cos(2\phi^{2}_{pol})} &  =  \notag \\ & \hspace{-2.5cm} \frac{cos(2\phi^{1}_{pol})z_{x} + sin(2\phi^{1}_{pol})z_{y}}{cos(2\phi^{2}_{pol})z_{x} + sin(2\phi^{2}_{pol})z_{y}}
	\label{eq:20}
	\end{align}
	Cross multiplying and rearranging the Equation \ref{eq:20} gives us the following.
	\begin{align}
	\centering
	\bigg(\Big(I(\phi^{1}_{pol}) - A_{d} & + B_{d}cos(2\phi^{1}_{pol})\Big)cos(2\phi^{2}_{pol}) \notag \\ 
	& \hspace{-2.5cm} - \Big(I(\phi^{1}_{pol}) - A_{d} + B_{d}cos(2\phi^{2}_{pol})\Big)cos(2\phi^{1}_{pol})\bigg)z_{x} \notag \\
	& \hspace{-2.5cm} + \bigg(\Big(I(\phi^{1}_{pol}) - A_{d} + B_{d}cos(2\phi^{1}_{pol})\Big)sin(2\phi^{2}_{pol}) \notag \\ 
	& \hspace{-2.5cm} - \Big(I(\phi^{1}_{pol}) - A_{d} + B_{d}cos(2\phi^{2}_{pol})\Big)sin(2\phi^{1}_{pol})\bigg)z_{y}  = 0 
	\label{eq:21}
	\end{align}
	Evaluating Equation \ref{eq:21} at $\phi^{1}_{pol} = 0$ and $\phi^{2}_{pol} = \frac{\pi}{4}$, we get the final form as follows.
	\begin{equation}
	\centering
	F_{d}z_{x} + G_{d}z_{y} = 0
	\label{eq:22}
	\end{equation}
	Here, $F_{d} = \big(-I(\frac{\pi}{4}) + A_{d}\big)$ and $G_{d} = \big(I(0) - A_{d} + B_{d}\big)$ are the components of the bi-dimensional vector field $v = (F,G)^{T}$ characterizing the level set in the differential formulation $v^{T}\nabla z = 0$, as per Equation \ref{eq:21}.
	
	\subsection*{6.2 Specular Polarization}
	We need to account for a $\frac{\pi}{2}$ phase shift for specular polarisation. The bi-dimensional vector field $v$ describing the level-set at a specular pixel has orthogonal direction to those at the diffuse pixel, accounting for the inherent $\pi$-periodic ambiguity in the azimuth angle represented by the phase angle $\phi$ \cite{logothetis2019differential, mecca2017differential} such that the following holds.
	\begin{align}
	\centering
	- \bigg(\Big(I(\phi^{1}_{pol}) - A_{s} & + B_{s}cos(2\phi^{1}_{pol})\Big)sin(2\phi^{2}_{pol}) \notag \\ 
	& \hspace{-2.5cm} + \Big(I(\phi^{1}_{pol}) - A_{s} + B_{s}cos(2\phi^{2}_{pol})\Big)sin(2\phi^{1}_{pol})\bigg)z_{x} \notag \\
	& \hspace{-2.5cm} + \bigg(\Big(I(\phi^{1}_{pol}) - A_{s} + B_{s}cos(2\phi^{1}_{pol})\Big)cos(2\phi^{2}_{pol}) \notag \\ 
	& \hspace{-2.5cm} - \Big(I(\phi^{1}_{pol}) - A_{s} + B_{s}cos(2\phi^{2}_{pol})\Big)cos(2\phi^{1}_{pol})\bigg)z_{y} = 0
	\label{eq:23}
	\end{align}
	Again, evaluating Equation \ref{eq:23} at $\phi^{1}_{pol} = 0$ and $\phi^{2}_{pol}=\frac{\pi}{4}$, we get the following constraint.
	\begin{equation}
	\centering
	-G_{s}z_{x} + F_{s}z_{y} = 0
	\label{eq:24}
	\end{equation}
	Here, $F_{s} = \big(-I(\frac{\pi}{4}) + A_{s}\big)$ and $G_{s} = \big(I(0) - A_{s} + B_{s}\big)$. We use Equation \ref{eq:22} and \ref{eq:24} as the constraints over diffuse and specular regions, as described in the main paper.

 \newpage
	\bibliographystyle{eg-alpha-doi}
	
	\bibliography{egbib}

\newcommand{\etalchar}[1]{$^{#1}$}
\begin{thebibliography}{\uppercase{MEMF12}}

\bibitem[AH06]{atkinson2006recovery}
\textsc{Atkinson G.~A., Hancock E.~R.}:
\newblock Recovery of surface orientation from diffuse polarization.
\newblock \emph{IEEE transactions on image processing 15}, 6 (2006), 1653--1664.

\bibitem[AH07]{atkinson2007shape}
\textsc{Atkinson G.~A., Hancock E.~R.}:
\newblock Shape estimation using polarization and shading from two views.
\newblock \emph{IEEE transactions on pattern analysis and machine intelligence 29}, 11 (2007), 2001--2017.

\bibitem[Atk17]{atkinson2017polarisation}
\textsc{Atkinson G.~A.}:
\newblock Polarisation photometric stereo.
\newblock \emph{Computer Vision and Image Understanding 160} (2017), 158--167.

\bibitem[BGW{\etalchar{*}}20]{ba2020deep}
\textsc{Ba Y., Gilbert A., Wang F., Yang J., Chen R., Wang Y., Yan L., Shi B., Kadambi A.}:
\newblock Deep shape from polarization.
\newblock In \emph{European Conference on Computer Vision} (2020), Springer, pp.~554--571.

\bibitem[BJTK18]{baek2018simultaneous}
\textsc{Baek S.-H., Jeon D.~S., Tong X., Kim M.~H.}:
\newblock Simultaneous acquisition of polarimetric svbrdf and normals.
\newblock \emph{ACM Trans. Graph. 37}, 6 (2018), 268--1.

\bibitem[BVM17]{berger2017depth}
\textsc{Berger K., Voorhies R., Matthies L.~H.}:
\newblock Depth from stereo polarization in specular scenes for urban robotics.
\newblock In \emph{2017 IEEE international conference on robotics and automation (ICRA)} (2017), IEEE, pp.~1966--1973.

\bibitem[CGS{\etalchar{*}}17]{cui2017polarimetric}
\textsc{Cui Z., Gu J., Shi B., Tan P., Kautz J.}:
\newblock Polarimetric multi-view stereo.
\newblock In \emph{Proceedings of the IEEE conference on computer vision and pattern recognition} (2017), pp.~1558--1567.

\bibitem[CLY{\etalchar{*}}21]{chen2021transunet}
\textsc{Chen J., Lu Y., Yu Q., Luo X., Adeli E., Wang Y., Lu L., Yuille A.~L., Zhou Y.}:
\newblock Transunet: Transformers make strong encoders for medical image segmentation.
\newblock \emph{arXiv preprint arXiv:2102.04306} (2021).

\bibitem[Col05]{collett2005field}
\textsc{Collett E.}:
\newblock Field guide to polarization.
\newblock Spie Bellingham, WA.

\bibitem[DJZ{\etalchar{*}}21]{ding2021polarimetric}
\textsc{Ding Y., Ji Y., Zhou M., Kang S.~B., Ye J.}:
\newblock Polarimetric helmholtz stereopsis.
\newblock In \emph{Proceedings of the IEEE/CVF International Conference on Computer Vision} (2021), pp.~5037--5046.

\bibitem[DLG21]{deschaintre2021deep}
\textsc{Deschaintre V., Lin Y., Ghosh A.}:
\newblock Deep polarization imaging for 3d shape and svbrdf acquisition.
\newblock In \emph{Proceedings of the IEEE/CVF Conference on Computer Vision and Pattern Recognition} (2021), pp.~15567--15576.

\bibitem[DZV22]{dave2022pandora}
\textsc{Dave A., Zhao Y., Veeraraghavan A.}:
\newblock Pandora: Polarization-aided neural decomposition of radiance.
\newblock \emph{arXiv preprint arXiv:2203.13458} (2022).

\bibitem[FKNN21]{fukao2021polarimetric}
\textsc{Fukao Y., Kawahara R., Nobuhara S., Nishino K.}:
\newblock Polarimetric normal stereo.
\newblock In \emph{Proceedings of the IEEE/CVF Conference on Computer Vision and Pattern Recognition} (2021), pp.~682--690.

\bibitem[GCP{\etalchar{*}}10]{ghosh2010circularly}
\textsc{Ghosh A., Chen T., Peers P., Wilson C.~A., Debevec P.}:
\newblock Circularly polarized spherical illumination reflectometry.
\newblock In \emph{ACM SIGGRAPH Asia 2010 papers}. 2010, pp.~1--12.

\bibitem[GFT{\etalchar{*}}11]{ghosh2011multiview}
\textsc{Ghosh A., Fyffe G., Tunwattanapong B., Busch J., Yu X., Debevec P.}:
\newblock Multiview face capture using polarized spherical gradient illumination.
\newblock In \emph{Proceedings of the 2011 SIGGRAPH Asia Conference} (2011), pp.~1--10.

\bibitem[GPDG12]{guarnera2012estimating}
\textsc{Guarnera G.~C., Peers P., Debevec P., Ghosh A.}:
\newblock Estimating surface normals from spherical stokes reflectance fields.
\newblock In \emph{Computer Vision--ECCV 2012. Workshops and Demonstrations: Florence, Italy, October 7-13, 2012, Proceedings, Part II 12} (2012), Springer, pp.~340--349.

\bibitem[HRKH10]{huynh2010shape}
\textsc{Huynh C.~P., Robles-Kelly A., Hancock E.}:
\newblock Shape and refractive index recovery from single-view polarisation images.
\newblock In \emph{2010 IEEE Computer Society Conference on Computer Vision and Pattern Recognition} (2010), IEEE, pp.~1229--1236.

\bibitem[HRKH13]{huynh2013shape}
\textsc{Huynh C.~P., Robles-Kelly A., Hancock E.~R.}:
\newblock Shape and refractive index from single-view spectro-polarimetric images.
\newblock \emph{International journal of computer vision 101}, 1 (2013), 64--94.

\bibitem[HSL{\etalchar{*}}16]{huang2016deep}
\textsc{Huang G., Sun Y., Liu Z., Sedra D., Weinberger K.~Q.}:
\newblock Deep networks with stochastic depth.
\newblock In \emph{Computer Vision--ECCV 2016: 14th European Conference, Amsterdam, The Netherlands, October 11--14, 2016, Proceedings, Part IV 14} (2016), Springer, pp.~646--661.

\bibitem[KB14]{kingma2014adam}
\textsc{Kingma D.~P., Ba J.}:
\newblock Adam: A method for stochastic optimization.
\newblock \emph{arXiv preprint arXiv:1412.6980} (2014).

\bibitem[KOS{\etalchar{*}}20]{kondo2020accurate}
\textsc{Kondo Y., Ono T., Sun L., Hirasawa Y., Murayama J.}:
\newblock Accurate polarimetric brdf for real polarization scene rendering.
\newblock In \emph{European Conference on Computer Vision} (2020), Springer, pp.~220--236.

\bibitem[KPKO23]{kajiyama2023separating}
\textsc{Kajiyama S., Piao T., Kawahara R., Okabe T.}:
\newblock Separating partially-polarized diffuse and specular reflection components under unpolarized light sources.
\newblock In \emph{Proceedings of the IEEE/CVF Winter Conference on Applications of Computer Vision} (2023), pp.~2549--2558.

\bibitem[KTSR15]{kadambi2015polarized}
\textsc{Kadambi A., Taamazyan V., Shi B., Raskar R.}:
\newblock Polarized 3d: High-quality depth sensing with polarization cues.
\newblock In \emph{Proceedings of the IEEE International Conference on Computer Vision} (2015), pp.~3370--3378.

\bibitem[LLK{\etalchar{*}}02]{lin2002diffuse}
\textsc{Lin S., Li Y., Kang S.~B., Tong X., Shum H.-Y.}:
\newblock Diffuse-specular separation and depth recovery from image sequences.
\newblock In \emph{European conference on computer vision} (2002), Springer, pp.~210--224.

\bibitem[LMSC19]{logothetis2019differential}
\textsc{Logothetis F., Mecca R., Sgallari F., Cipolla R.}:
\newblock A differential approach to shape from polarisation: A level-set characterisation.
\newblock \emph{International Journal of Computer Vision 127} (2019), 1680--1693.

\bibitem[LQX{\etalchar{*}}22]{lei2022shape}
\textsc{Lei C., Qi C., Xie J., Fan N., Koltun V., Chen Q.}:
\newblock Shape from polarization for complex scenes in the wild.
\newblock In \emph{Proceedings of the IEEE/CVF Conference on Computer Vision and Pattern Recognition} (2022), pp.~12632--12641.

\bibitem[MBMS23]{muglikar2023event}
\textsc{Muglikar M., Bauersfeld L., Moeys D.~P., Scaramuzza D.}:
\newblock Event-based shape from polarization.
\newblock \emph{arXiv preprint arXiv:2301.06855} (2023).

\bibitem[MCZ{\etalchar{*}}22]{mingqi2022transparent}
\textsc{Mingqi S., Chongkun X., Zhendong Y., Junnan H., Xueqian W.}:
\newblock Transparent shape from single polarization images.
\newblock \emph{arXiv preprint arXiv:2204.06331} (2022).

\bibitem[MEMF12]{mahmoud2012direct}
\textsc{Mahmoud A.~H., El-Melegy M.~T., Farag A.~A.}:
\newblock Direct method for shape recovery from polarization and shading.
\newblock In \emph{2012 19th IEEE International Conference on Image Processing} (2012), IEEE, pp.~1769--1772.

\bibitem[MFSG06]{morel2006active}
\textsc{Morel O., Ferraton M., Stolz C., Gorria P.}:
\newblock Active lighting applied to shape from polarization.
\newblock In \emph{2006 International Conference on Image Processing} (2006), IEEE, pp.~2181--2184.

\bibitem[MHP{\etalchar{*}}07]{ma2007rapid}
\textsc{Ma W.-C., Hawkins T., Peers P., Chabert C.-F., Weiss M., Debevec P.~E., et~al.}:
\newblock Rapid acquisition of specular and diffuse normal maps from polarized spherical gradient illumination.
\newblock \emph{Rendering Techniques 2007}, 9 (2007), 10.

\bibitem[MLC17]{mecca2017differential}
\textsc{Mecca R., Logothetis F., Cipolla R.}:
\newblock A differential approach to shape from polarization.

\bibitem[MSB{\etalchar{*}}16]{miyazaki2016surface}
\textsc{Miyazaki D., Shigetomi T., Baba M., Furukawa R., Hiura S., Asada N.}:
\newblock Surface normal estimation of black specular objects from multiview polarization images.
\newblock \emph{Optical Engineering 56}, 4 (2016), 041303.

\bibitem[MTHI03]{miyazaki2003polarization}
\textsc{Miyazaki D., Tan R.~T., Hara K., Ikeuchi K.}:
\newblock Polarization-based inverse rendering from a single view.
\newblock In \emph{Computer Vision, IEEE International Conference on} (2003), vol.~3, IEEE Computer Society, pp.~982--982.

\bibitem[NFB97]{nayar1997separation}
\textsc{Nayar S.~K., Fang X.-S., Boult T.}:
\newblock Separation of reflection components using color and polarization.
\newblock \emph{International Journal of Computer Vision 21}, 3 (1997), 163--186.

\bibitem[NZI01]{nishino2001determining}
\textsc{Nishino K., Zhang Z., Ikeuchi K.}:
\newblock Determining reflectance parameters and illumination distribution from a sparse set of images for view-dependent image synthesis.
\newblock In \emph{Proceedings Eighth IEEE international conference on computer vision. ICCV 2001} (2001), vol.~1, IEEE, pp.~599--606.

\bibitem[PF05]{prados2005shape}
\textsc{Prados E., Faugeras O.}:
\newblock Shape from shading: a well-posed problem?
\newblock In \emph{2005 IEEE computer society conference on computer vision and pattern recognition (CVPR'05)} (2005), vol.~2, IEEE, pp.~870--877.

\bibitem[PGC{\etalchar{*}}17]{paszke2017automatic}
\textsc{Paszke A., Gross S., Chintala S., Chanan G., Yang E., DeVito Z., Lin Z., Desmaison A., Antiga L., Lerer A.}:
\newblock Automatic differentiation in pytorch.

\bibitem[PLWZ19]{park2019semantic}
\textsc{Park T., Liu M.-Y., Wang T.-C., Zhu J.-Y.}:
\newblock Semantic image synthesis with spatially-adaptive normalization.
\newblock In \emph{Proceedings of the IEEE/CVF conference on computer vision and pattern recognition} (2019), pp.~2337--2346.

\bibitem[RC01]{rahmann2001reconstruction}
\textsc{Rahmann S., Canterakis N.}:
\newblock Reconstruction of specular surfaces using polarization imaging.
\newblock In \emph{Proceedings of the 2001 IEEE Computer Society Conference on Computer Vision and Pattern Recognition. CVPR 2001} (2001), vol.~1, IEEE, pp.~I--I.

\bibitem[Sch11]{schechner2011inversion}
\textsc{Schechner Y.~Y.}:
\newblock Inversion by p 4: polarization-picture post-processing.
\newblock \emph{Philosophical Transactions of the Royal Society B: Biological Sciences 366}, 1565 (2011), 638--648.

\bibitem[Sch15]{schechner2015self}
\textsc{Schechner Y.~Y.}:
\newblock Self-calibrating imaging polarimetry.
\newblock In \emph{2015 IEEE International Conference on Computational Photography (ICCP)} (2015), IEEE, pp.~1--10.

\bibitem[SDMF12]{shabayek2012vision}
\textsc{Shabayek A. E.~R., Demonceaux C., Morel O., Fofi D.}:
\newblock Vision based uav attitude estimation: Progress and insights.
\newblock \emph{Journal of Intelligent \& Robotic Systems 65}, 1 (2012), 295--308.

\bibitem[SRT18]{smith2018height}
\textsc{Smith W.~A., Ramamoorthi R., Tozza S.}:
\newblock Height-from-polarisation with unknown lighting or albedo.
\newblock \emph{IEEE transactions on pattern analysis and machine intelligence 41}, 12 (2018), 2875--2888.

\bibitem[SYC{\etalchar{*}}20]{shi2020recent}
\textsc{Shi B., Yang J., Chen J., Zhang R., Chen R.}:
\newblock Recent progress in shape from polarization.
\newblock \emph{Advances in Photometric 3D-Reconstruction} (2020), 177--203.

\bibitem[TI08]{tan2008separating}
\textsc{Tan R.~T., Ikeuchi K.}:
\newblock Separating reflection components of textured surfaces using a single image.
\newblock In \emph{Digitally Archiving Cultural Objects}. Springer, 2008, pp.~353--384.

\bibitem[TKR16]{taamazyan2016shape}
\textsc{Taamazyan V., Kadambi A., Raskar R.}:
\newblock Shape from mixed polarization.
\newblock \emph{arXiv preprint arXiv:1605.02066} (2016).

\bibitem[TZS{\etalchar{*}}21]{tozza2021uncalibrated}
\textsc{Tozza S., Zhu D., Smith W.~A., Ramamoorthi R., Hancock E.~R.}:
\newblock Uncalibrated, two source photo-polarimetric stereo.
\newblock \emph{IEEE Transactions on Pattern Analysis and Machine Intelligence 44}, 9 (2021), 5747--5760.

\bibitem[WB93]{wolff1993constraining}
\textsc{Wolff L.~B., Boult T.~E.}:
\newblock Constraining object features using a polarization reflectance model.
\newblock \emph{Phys. Based Vis. Princ. Pract. Radiom 1} (1993), 167.

\bibitem[Woo80]{woodham1980photometric}
\textsc{Woodham R.~J.}:
\newblock Photometric method for determining surface orientation from multiple images.
\newblock \emph{Optical engineering 19}, 1 (1980), 139--144.

\bibitem[YTD{\etalchar{*}}21]{yang2021transformer}
\textsc{Yang G., Tang H., Ding M., Sebe N., Ricci E.}:
\newblock Transformer-based attention networks for continuous pixel-wise prediction.
\newblock In \emph{Proceedings of the IEEE/CVF International Conference on Computer Vision} (2021), pp.~16269--16279.

\bibitem[YZS17]{yu2017shape}
\textsc{Yu Y., Zhu D., Smith W.~A.}:
\newblock Shape-from-polarisation: a nonlinear least squares approach.
\newblock In \emph{Proceedings of the IEEE International Conference on Computer Vision Workshops} (2017), pp.~2969--2976.

\bibitem[ZS19]{zhu2019depth}
\textsc{Zhu D., Smith W.~A.}:
\newblock Depth from a polarisation+ rgb stereo pair.
\newblock In \emph{Proceedings of the IEEE/CVF Conference on Computer Vision and Pattern Recognition} (2019), pp.~7586--7595.

\end{thebibliography}
	
	\newpage
	
\end{document}